\documentclass[10pt,twocolumn,letterpaper]{article}

\usepackage{cvpr}              

\usepackage{graphicx}
\usepackage{amsmath}
\usepackage{amssymb}
\usepackage{booktabs}
\usepackage[euler]{textgreek}

\usepackage{graphbox}
\usepackage{paralist}

\usepackage{makecell}
\usepackage{tabularx}
\usepackage{multirow}
\usepackage{multicol}

\usepackage[T1]{fontenc}    
\usepackage{amsfonts}       
\usepackage{pifont}
\usepackage{dsfont}

\usepackage{algorithm}
\usepackage{algorithmic}

\usepackage{color}
\usepackage[dvipsnames,svgnames]{xcolor}
\usepackage{colortbl}
\definecolor{Gray}{gray}{0.85}
\definecolor{LightCyan}{rgb}{0.88,1,1}

\usepackage[utf8]{inputenc} 
\usepackage{url}            
\usepackage{mathtools}
\usepackage{nicefrac}       
\usepackage{ragged2e}
\usepackage{mwe}

\usepackage[normalem]{ulem}

\usepackage{url}
\usepackage[pagebackref,breaklinks,colorlinks,citecolor=MyGreen]{hyperref}

\definecolor{darkergreen}{RGB}{21, 152, 56}
\definecolor{red2}{RGB}{252, 54, 65}
\newcommand{\cmark}{\textcolor{darkergreen}{\ding{51}}}%
\newcommand{\xmark}{\textcolor{red2}{\ding{55}}}%

\usepackage[capitalize]{cleveref}
\crefname{section}{Sec.}{Secs.}
\Crefname{section}{Section}{Sections}
\Crefname{table}{Table}{Tables}
\crefname{table}{Tab.}{Tabs.}

\usepackage[accsupp]{axessibility}  

\def\cvprPaperID{4973} 
\def\confName{CVPR}
\def\confYear{2023}

\begin{document}


\long\def\ignorethis#1{}



\newcommand{\img}[1]{\mathbf{I}_{\text{#1}}}
\newcommand{\paren}[1]{\left( #1 \right)}
\newcommand{\bparen}[1]{\left[ #1 \right]}
\newcommand{\feature}[1]{\phi \paren{#1}}
\newcommand{\normtwo}[1]{\lVert #1 \rVert_2^2}
\newcommand{\normone}[1]{\left\lVert #1 \right\rVert_1}

\newcommand{\Paragraph}[1]{\vspace{1mm}\noindent\textbf{#1}}

\newcommand{\figref}[1]{Figure~\ref{fig:#1}}
\newcommand{\tabref}[1]{Table~\ref{tab:#1}} 
\newcommand{\itmref}[1]{[\ref{itm:#1}]}     
\newcommand{\eqnref}[1]{\eqref{eq:#1}}
\newcommand{\secref}[1]{Section~\ref{sec:#1}}
\newcommand{\eqmain}[1]{(\textcolor{blue}{#1})}
\newcommand{\fakeref}[1]{\textcolor{MyGreen}{#1}}
\newcommand{\fakeeqref}[1]{\textcolor{MyGreen}{(#1)}}

\newcommand{\mb}[1]{\mathbf{#1}}
\newcommand{\bs}[1]{\boldsymbol{#1}}
\newcommand{\n}{\mbox{\qquad}}              
\newcommand{\red}[1]{{\color{red}#1}}

\newcommand{\ignore}[1]{}   
\newcommand{\cmt}[1]{\begin{sloppypar}\large\textcolor{red}{#1}\end{sloppypar}}

\newcommand{\TODO}[1]{\textcolor{red}{[TODO]\{#1\}}}
\newcommand{\todo}[1]{\textcolor{red}{#1}}
\newcommand{\torevise}[1]{\textcolor{blue}{#1}}
\newcommand{\revise}[1]{\textcolor{blue}{#1}}
\newcommand{\copied}[1]{\textcolor{red}{[COPIED: #1]}}

\newcommand{\needref}{[\textcolor{blue}{put}, \textcolor{blue}{some}, \textcolor{blue}{references}]}

\newcommand{\tabspace}{\vspace{-2mm}}
\newcommand{\tabxspace}{\vspace{-4mm}}
\newcommand{\figspace}{\vspace{-2mm}}
\newcommand{\figxspace}{\vspace{-3mm}}
\newcommand{\best}[1]{{\textcolor{red}{\textbf{#1}}}}
\newcommand{\secondbest}[1]{{\textcolor{blue}{\underline{#1}}}}

\newcommand{\ts}{\textsuperscript}

\newcommand{\set}[1]{\{#1\}}


\def\PE{\Phi}

\newcommand{\mpage}[2]
{
\begin{minipage}{#1\linewidth}\centering
#2
\end{minipage}
}

\newcommand{\replace}[2]{\textcolor{red}{\sout{#1}} \textcolor{blue}{#2}}
\newcommand{\fix}[1]{\textcolor{red}{#1}}
\newcommand{\proposed}{UDMM}

\newcommand{\bsd}{MSD}

\newcommand{\blurhigh}{\mathbf{I}_{\hat{\text{B}}}}
\newcommand{\blurlow}{\mathbf{I}_{\hat{\text{B}},\hat{\text{LR}}}}

\definecolor{MyGreen}{cmyk}{100, 0, 100, 0}

\newcommand{\citenumber}[1]{[\textcolor{MyGreen}{#1}]}
\newcommand{\refnumber}[1]{\textcolor{red}{#1}}

\newcolumntype{L}[1]{>{\raggedright\let\newline\\\arraybackslash\hspace{0pt}}m{#1}}
\newcolumntype{C}[1]{>{\centering\let\newline\\\arraybackslash\hspace{0pt}}m{#1}}
\newcolumntype{R}[1]{>{\raggedleft\let\newline\\\arraybackslash\hspace{0pt}}m{#1}}

\newcommand{\jaeha}[1]{{\textcolor{BurntOrange}{\textbf{Jaeha: }#1}}}
\newcommand{\joonkyu}[1]{{\textcolor{blue}{\textbf{Joonkyu: }#1}}}
\newcommand{\younguk}[1]{{\textcolor{ProcessBlue}{\textbf{Younguk: }#1}}}


\title{Recovering 3D Hand Mesh Sequence from a Single Blurry Image: \\ A New Dataset and Temporal Unfolding}

\author{$\text{Yeonguk Oh}^{1\ast}$\quad $\text{JoonKyu Park}^{1\ast}$\quad $\text{Jaeha Kim}^{1\ast}$\quad $\text{Gyeongsik Moon}^{3}$\quad $\text{Kyoung Mu Lee}^{1,2}$\\
$^{1}$Dept. of ECE\&ASRI, $^{2}$IPAI, Seoul National University, Korea\\$^{3}$Meta Reality Labs Research\\
{\tt\small \{namepllet, jkpark0825\}@snu.ac.kr, jhkim97s2@gmail.com, mks0601@meta.com, kyoungmu@snu.ac.kr}
}
\maketitle
\def\thefootnote{*}\footnotetext{Authors contributed equally.}\def\thefootnote{\arabic{footnote}}

\begin{abstract}
Hands, one of the most dynamic parts of our body, suffer from blur due to their active movements.
However, previous 3D hand mesh recovery methods have mainly focused on sharp hand images rather than considering blur due to the absence of datasets providing blurry hand images.
We first present a novel dataset BlurHand, which contains blurry hand images with 3D groundtruths.
The BlurHand is constructed by synthesizing motion blur from sequential sharp hand images, imitating realistic and natural motion blurs.
In addition to the new dataset, we propose BlurHandNet, a baseline network for accurate 3D hand mesh recovery from a blurry hand image.
Our BlurHandNet unfolds a blurry input image to a 3D hand mesh sequence to utilize temporal information in the blurry input image, while previous works output a static single hand mesh.
We demonstrate the usefulness of BlurHand for the 3D hand mesh recovery from blurry images in our experiments.
The proposed BlurHandNet produces much more robust results on blurry images while generalizing well to in-the-wild images.
The training codes and BlurHand dataset are available at \href{https://github.com/JaehaKim97/BlurHand_RELEASE}{https://github.com/JaehaKim97/BlurHand\_RELEASE}.

\end{abstract}

\section{Introduction}

\begin{figure}[t]
\begin{center}
\vspace{2mm}
\begin{subfigure}{0.93\linewidth}
\centering
\includegraphics[width=1.\linewidth]{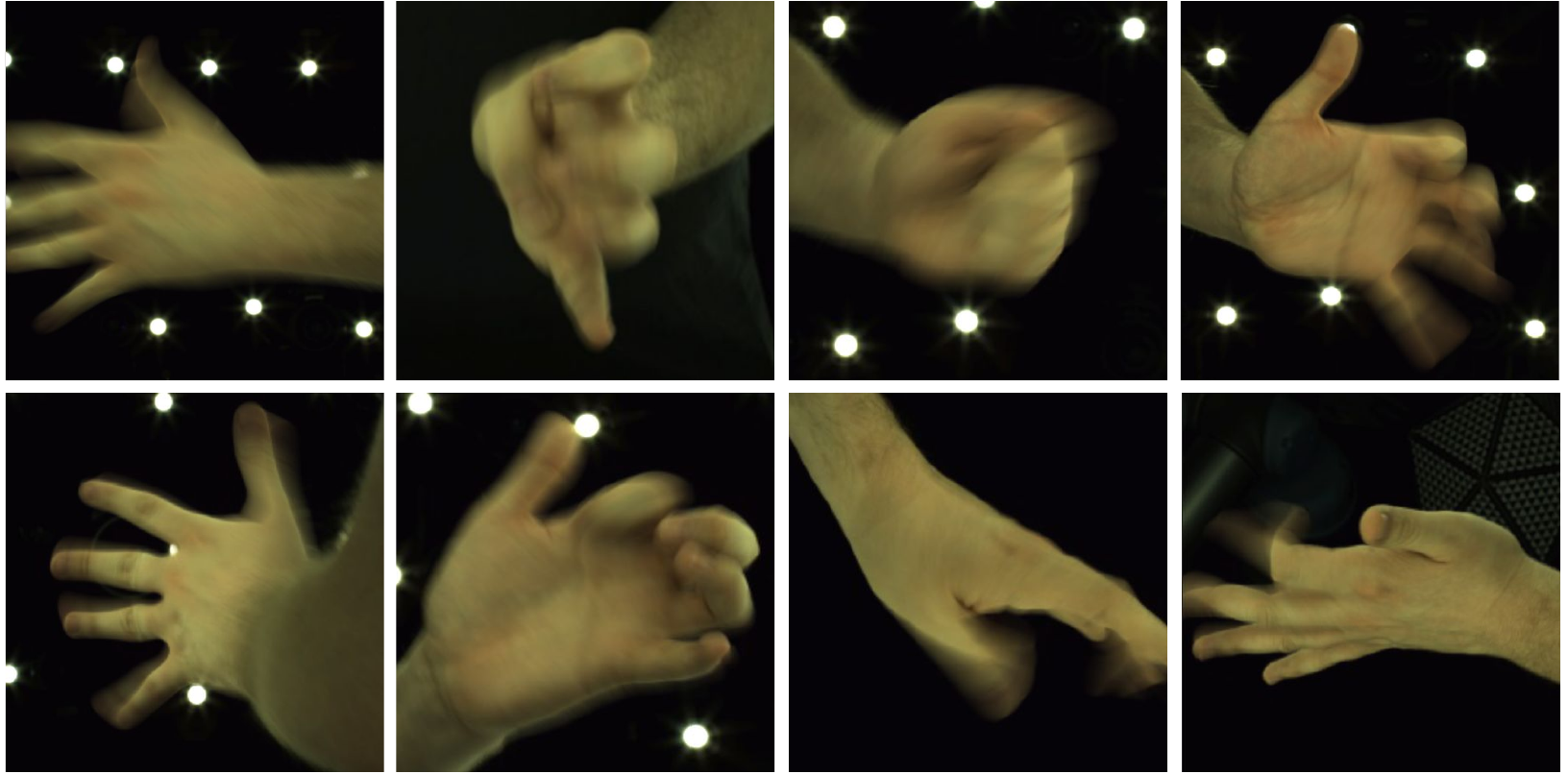}
\caption{\textbf{Examples of the presented BlurHand dataset.}
\label{fig1-a}}
\end{subfigure}
\\
\vspace{2mm}
\begin{subfigure}{0.93\linewidth}
\centering
\includegraphics[width=1.\linewidth]{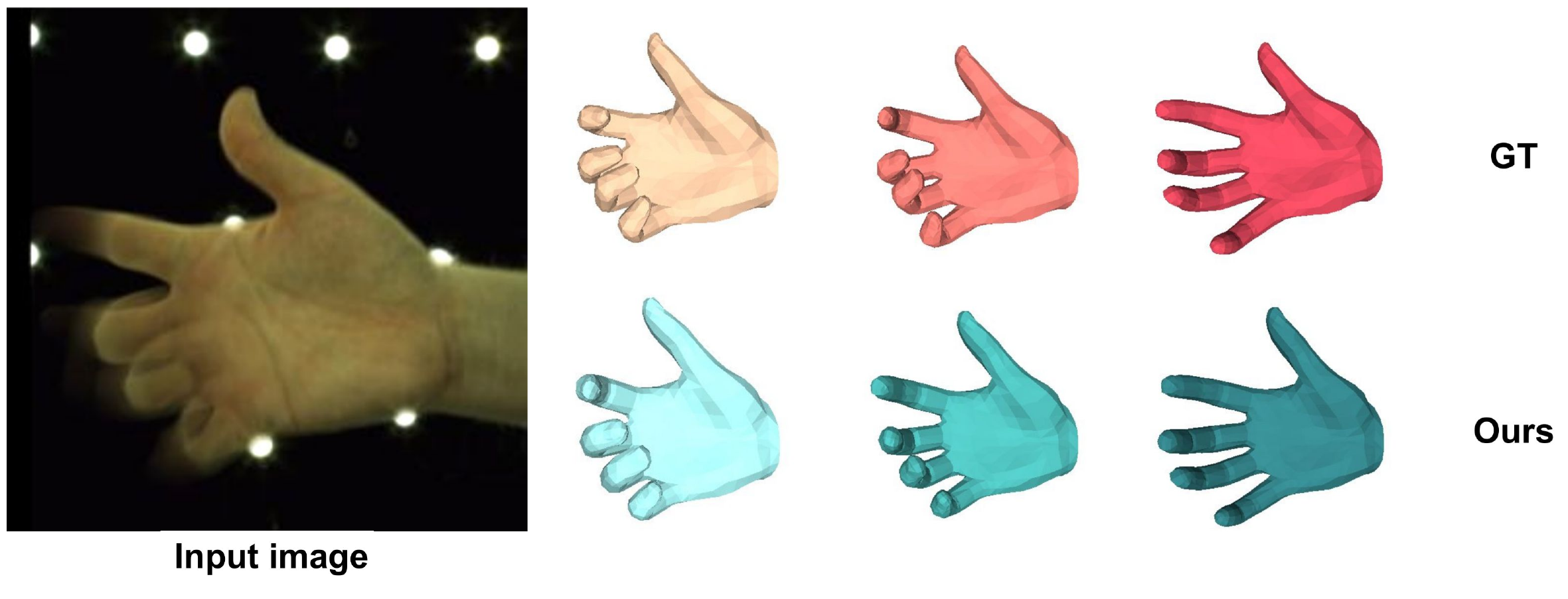}
\caption{\textbf{Illustration of the temporal unfolding.} \label{fig1-b}}
\end{subfigure}
\end{center}
\vspace{-6mm}
\caption{\textbf{Proposed BlurHand dataset and BlurHandNet.}
(a) We present a novel BlurHand dataset, providing natural blurry hand images with accurate 3D annotations.
(b) While most previous methods produce a single 3D hand mesh from a sharp image, our BlurHandNet unfolds the blurry hand image into three sequential hand meshes.}
\vspace{-3mm}
\label{fig:overview}
\end{figure}

Since hand images frequently contain blur when hands are moving, developing a blur-robust 3D hand mesh estimation framework is necessary.
As blur makes the boundary unclear and hard to recognize, it significantly degrades the performance of 3D hand mesh estimation and makes the task challenging.
Despite promising results of 3D hand mesh estimation from a single sharp image~\cite{choi2020pose2mesh, moon2020i2l, kulon2020weakly, lin2021end, lin2021mesh}, research on blurry hands is barely conducted.

A primary reason for such lack of consideration is the absence of datasets that consist of blurry hand images with accurate 3D groundtruth~(GT).
Capturing blurry hand datasets is greatly challenging.
The standard way of capturing markerless 3D hand datasets~\cite{hampali2020honnotate, moon2020interhand2,zimmermann2019freihand} consists of two stages: 1) obtaining multi-view 2D grounds (\eg, 2D joint coordinates and mask) manually~\cite{zimmermann2019freihand} or using estimators~\cite{wei2016convolutional,chen2017rethinking,li2019rethinking} and 2) triangulating the multi-view 2D grounds to the 3D space.
Here, manual annotations or estimators in the first stage are performed from images.
Hence, they become unreliable when the input image is blurry, which results in triangulation failure in the second stage.

Contemplating these limitations, we present the BlurHand, whose examples are shown in Figure~\ref{fig1-a}.  
Our BlurHand, the first blurry hand dataset, is synthesized from InterHand2.6M~\cite{moon2020interhand2}, which is a widely adopted video-based hand dataset with accurate 3D annotations.
Following state-of-the-art blur synthesis literature~\cite{Su_2017_CVPR,Nah_2017_CVPR,Nah_2019_ICCV_Workshops}, we approximate the blurry images by averaging the sequence of sharp hand frames. 
As such technique requires high frame rates of videos, we employ a widely used video interpolation method~\cite{Niklaus_ICCV_2017} to complement the low frame rate~(30 frames per second) of InterHand2.6M. 
We note that our synthetic blur dataset contains realistic and challenging blurry hands.

For a given blurry hand image, the most straightforward baseline is sequentially applying state-of-the-art deblurring methods~\cite{chen2022simple,zamir2022restormer,park2022pay,park2022recurrence} on blurry images and 3D hand mesh estimation networks~\cite{moon2020i2l,spurr2018cross,moon2022accurate} on the deblurred image.
However, such a simple baseline suffers from two limitations.
First, since hands contain challenging blur caused by complex articulations, even state-of-the-art deblurring methods could not completely deblur the image.
Therefore, the performance of the following 3D hand mesh estimation networks severely drops due to remaining blur artifacts.
Second, since conventional deblurring approaches only restore the sharp images corresponding to the middle of the motion, it limits the chance to make use of temporal information, which might be useful for 3D mesh estimation.
In other words, the deblurring process restricts networks from exploiting the motion information in blurry hand images.

To overcome the limitations, we propose BlurHandNet, which recovers a 3D hand mesh sequence from a single blurry image, as shown in Figure~\ref{fig1-b}.
Our BlurHandNet effectively incorporates useful temporal information from the blurry hand.
The main components of BlurHandNet are Unfolder and a kinematic temporal Transformer~(KTFormer).
Unfolder outputs hand features of three timesteps, \ie, middle and both ends of the motion~\cite{Jin_2018_CVPR, Purohit_2019_CVPR, PAN_2019_CVPR, Rozumnyi2021defmo}.
The Unfolder brings benefits to our method in two aspects.
First, Unfolder enables the proposed BlurHandNet to output not only 3D mesh in the middle of the motion but also 3D meshes at both ends of the motion, providing more informative results related to motion.
We note that this property is especially beneficial for the hands, where the motion has high practical value in various hand-related works.
For example, understanding hand motion is essential in the domain of sign language~\cite{boulares2019sign, Rodríguez2020signdataset} and hand gestures~\cite{Sugimura2016motiongesture}, where the movement itself represents meaning.
Second, extracting features from multiple time steps enables the following modules to employ temporal information effectively.
Since hand features in each time step are highly correlated, exploiting temporal information benefits reconstructing more accurate 3D hand mesh estimation.

To effectively incorporate temporal hand features from the Unfolder, we propose KTFormer as the following module.
The KTFormer takes temporal hand features as input and leverages self-attention to enhance the temporal hand features.
The KTFormer enables the proposed BlurHandNet to implicitly consider both the kinematic structure and temporal relationship between the hands in three timesteps.
The KTFormer brings significant performance gain when coupled with Unfolder, demonstrating that employing temporal information plays a key role in accurate 3D hand mesh estimation from blurry hand images.


With a combination of BlurHand and BlurHandNet, we first tackle 3D hand mesh recovery from blurry hand images.
We show that BlurHandNet produces robust results from blurry hands and further demonstrate that BlurHandNet generalizes well on in-the-wild blurry hand images by taking advantage of effective temporal modules and BlurHand.
As this problem is barely studied, we hope our work could provide useful insights into the following works.
We summarize our contributions as follows:
\begin{itemize}
    \item We present a novel blurry hand dataset, BlurHand, which contains natural blurry hand images with accurate 3D GTs. 
    \item We propose the BlurHandNet for accurate 3D hand mesh estimation from blurry hand images with novel temporal modules, Unfolder and KTFormer.
    \item We experimentally demonstrate that the proposed BlurHandNet achieves superior 3D hand mesh estimation performance on blurry hands. 
\end{itemize}

\section{Related works}

\noindent\textbf{3D hand mesh estimation.}
Since after the introduction of RGB-based hand benchmark datasets with accurate 3D annotations, \eg, Friehand~\cite{zimmermann2019freihand} and InterHand 2.6M~\cite{moon2020interhand2},
various monocular RGB-based 3D hand mesh estimation methods~\cite{moon2020i2l, choi2020pose2mesh, kulon2020weakly, lin2021end, lin2021mesh, park2022handoccnet,moon2022accurate} have been proposed.
Pose2Mesh~\cite{choi2020pose2mesh} proposed a framework that reconstructs 3D mesh from the skeleton pose based on graph convolutional networks. Kulon~\etal~\cite{kulon2020weakly} utilized encoder-decoder architecture with a spiral operator to regress the 3D hand mesh.
I2L-MeshNet~\cite{moon2020i2l} utilized a 1D heatmap for each mesh vertex to model the uncertainty and preserve the spatial structure.
I2UV-HandNet~\cite{chen2021i2uv} proposed UV-based 3D hand shape representation and 3D hand super-resolution module to obtain high-fidelity hand meshes.
Pose2Pose~\cite{moon2022accurate} introduced joint features and proposed a 3D positional pose-guided 3D rotational pose prediction framework.
More recently, LISA~\cite{corona2022lisa} captured precise hand shape and appearance while providing dense surface correspondence, allowing for easy animation of the outputs.
SeqHAND~\cite{yang2020seqhand} incorporated synthetic datasets to train a recurrent framework with temporal movement information and consistency constraints, improving general pose estimations.
Meng~\etal~\cite{meng20223d} decomposed the 3D hand pose estimation task and used the HDR framework to handle occlusion.

After the success of the attention-based mechanism, Transformer~\cite{vaswani2017attention} has been adopted to recover more accurate 3D hand meshes.
METRO~\cite{lin2021end} and MeshGraphormer~\cite{lin2021mesh} proposed Transformer-based architecture, which models vertex-vertex and vertex-joint interactions.
Liu~\etal~\cite{liu2022spatial} utilizes spatial-temporal parallel Transformer to model inter-correlation between arm and hand.
HandOccNet~\cite{park2022handoccnet} proposed a Transformer-based feature injection mechanism to robustly reconstruct 3D hand mesh when occlusions are severe.
Although the above methods showed promising results for the sharp hand images, none of them carefully considered the hand with blur scenario.
As the lack of an appropriate dataset is the main reason for the less consideration, we present BlurHand.
Furthermore, we introduce a baseline network, BlurHandNet, which consists of a temporal unfolding module and kinematic temporal Transformer.


\noindent\textbf{Restoring the motion from a single blurry image.}
%
Rather than reconstructing only a single sharp image in the middle of the motion, recent deblurring methods~\cite{Jin_2018_CVPR, Purohit_2019_CVPR, Zhang_2020_ACMMM, PAN_2019_CVPR, Argaw_2021_CVPRW} have witnessed predicting the sequence of sharp frames from a single blurry image, which constructs the blurry input image.
Such a sequence of sharp frames can provide useful temporal information.
Jin~\etal~\cite{Jin_2018_CVPR} proposed temporal order invariant loss to overcome the temporal order ambiguity problem.
Purohit~\etal~\cite{Purohit_2019_CVPR} proposed an RNN-based solution without constraining the number of frames in sequence.
%
Argaw~\etal~\cite{Argaw_2021_CVPRW} proposed an encoder-decoder-based spatial Transformer network with regularizing terms.
%
Unlike previous methods that proposed to restore a single sharp image, our BlurHandNet aims to recover 3D hand mesh sequences from a single blurry image.

\section{BlurHand dataset}
\label{sec:blurhand_dataset}
%
%
%
%

\begin{figure}[t]
\begin{center}
\subfloat{\includegraphics[width=0.95\linewidth]{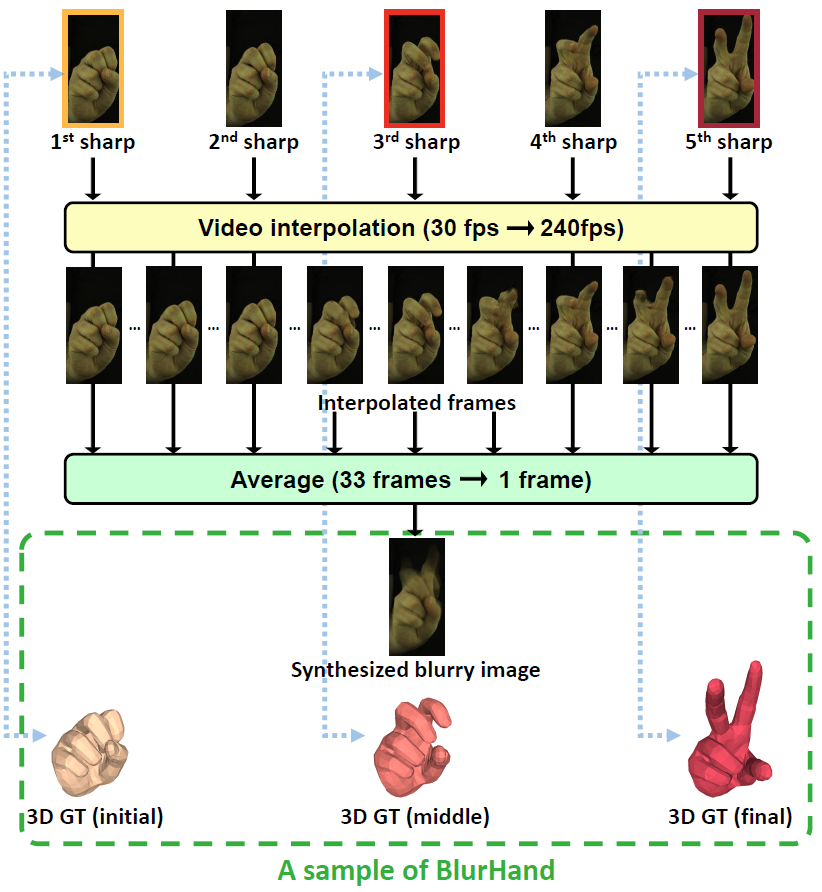}}\end{center}
\vspace{-7mm}
\caption{\textbf{Pipeline for constructing our BlurHand dataset.}
We synthesize the blurry hand from five sequential sharp hand frames by adopting video interpolation and averaging them.
%
}
\vspace{-4mm}
\label{fig:manufacture_diagram}
\end{figure}

%
\figref{manufacture_diagram} shows the overall pipeline for constructing our BlurHand.
Our BlurHand dataset is synthesized using 30 frames per second (fps) version of InterHand2.6M~\cite{moon2020interhand2}, which contains large-scale hand videos with diverse poses.
%
We first apply a video interpolation method~\cite{Niklaus_ICCV_2017} to increase 30 fps videos to 240 fps ones.
Then, a single blurry hand image is synthesized by averaging 33 sequential frames, which are interpolated from 5 sharp sequential frames, following the conventional deblurring dataset manufacture~\cite{Nah_2017_CVPR, zhou2019davanet, HA2019hide}.
We note that video interpolation is necessary when synthesizing blurs, as averaging frames from a low frame rate induces unnatural artifacts such as spikes and steps~\cite{Nah_2019_CVPR_Workshops_REDS}.
%
For each synthesized blurry image, 3D GTs of \emph{1st}, \emph{3rd}, and \emph{5th} sharp frames from InterHand2.6M 30fps are assigned as 3D GTs of initial, middle, and final, respectively.
%
%
%
In the end, the presented BlurHand consists of 121,839 training and 34,057 test samples containing single and interacting blurry hand images.
%
%
During the synthesis of blurry frames, we skip the frames if two neighboring frames are not available, and further adopt camera view sampling to mitigate the redundancy of samples.
%
We report sample statistics of the BlurHand in the supplementary materials.

\begin{figure*}[t]
\begin{center}
\includegraphics[width=0.94\linewidth]{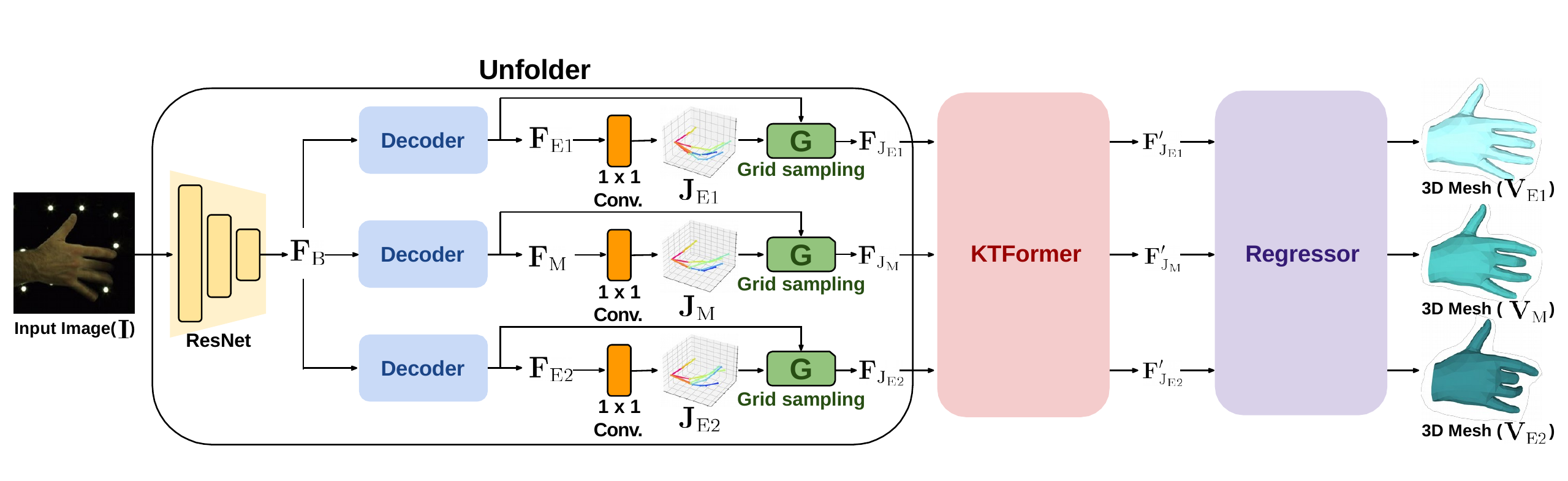}
\end{center}
\vspace{-9mm}
\caption{\textbf{Overall architecture of BlurHandNet.} BlurHandNet first unfolds input image~$\mathbf{I}$ into three temporal joint features~$\mathbf{F}_{{\text{J}}_{\text{E1}}}$, $\mathbf{F}_{{\text{J}}_{\text{E2}}}$, and $\mathbf{F}_{{\text{J}}_{\text{M}}}$.
The following kinematic temporal Transformer~(KTFormer) refines each joint feature by leveraging the attentive correlation between them.
Finally, Regressor produces MANO~\cite{romero2022embodied} parameters for each time step, resulting in temporal 3D hand meshes.
}
\vspace{-3mm}
\label{fig:model}
\end{figure*}

%
%
%
%

\section{BlurHandNet}
Figure~\ref{fig:model} shows the overall architecture of our BlurHandNet.
Our BlurHandNet, which consists of three modules; Unfolder, KTFormer, and Regressor, reconstructs sequential hand meshes from a single blurry hand image.
We describe the details of each module in the following sections.

\subsection{Unfolder}
\noindent\textbf{Unfolding a blurry hand image.}
Given a single RGB blurry hand image~$\mathbf{I} \in \mathbb{R}^{H \times W \times 3}$, Unfolder outputs feature maps and 3D joint coordinates of the three sequential hands, \ie, temporal unfolding, where each corresponds to the hand from both ends and the middle of the motion.
Here, $H=256$ and $W=256$ denote the height and width of the input image, respectively.
The temporal unfolding could extract useful temporal information from a single blurry image, and we note that effectively utilizing them is one of the core ideas of our methods.
To this end, we first feed the blurry hand image~$\mathbf{I}$ into ResNet50~\cite{he2016deep}, pre-trained on ImageNet~\cite{deng2009imagenet}, to extract the \emph{blurry} hand feature map~$\mathbf{F}_{\text{B}} \in \mathbb{R}^{\nicefrac{H}{32} \times \nicefrac{W}{32} \times C}$, where $C=2048$ denotes the channel dimension of $\mathbf{F}_{\text{B}}$.
Then, we predict three temporal features from a blurry hand feature~$\mathbf{F}_{\text{B}}$ through corresponding separate decoders, as shown in Figure~\ref{fig:model}.
As a result, we obtain three sequential hand features~$\mathbf{F}_{\text{E1}}$, $\mathbf{F}_{\text{E2}}$, and $\mathbf{F}_{\text{M}}$ with dimension~$\mathbb{R}^{h \times w \times c}$, where each corresponds to the hand at both ends and the middle of the motion.
Here, $h=\nicefrac{H}{4}$, $w=\nicefrac{W}{4}$, and $c=512$ denote the height, width, and channel dimension of each hand feature, respectively.

\begin{figure}[t]
\begin{center}
\subfloat[\label{fig:temporal_orderinga}]{\includegraphics[width=0.24\linewidth]{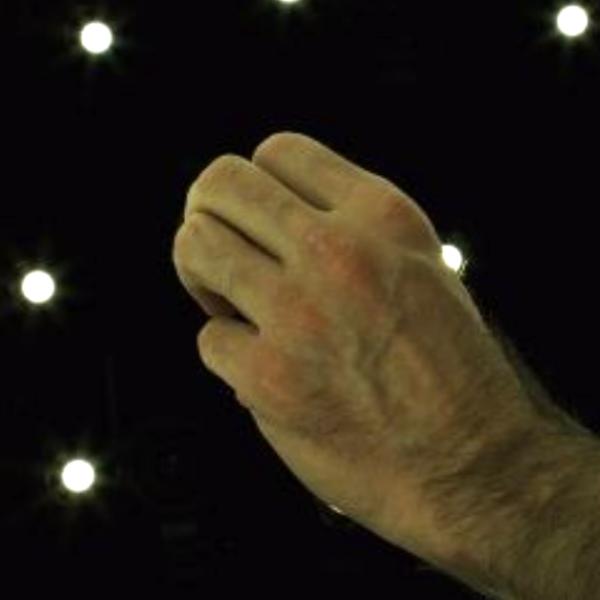}}
\hfill
\subfloat[\label{fig:temporal_orderingb}]{\includegraphics[width=0.24\linewidth]{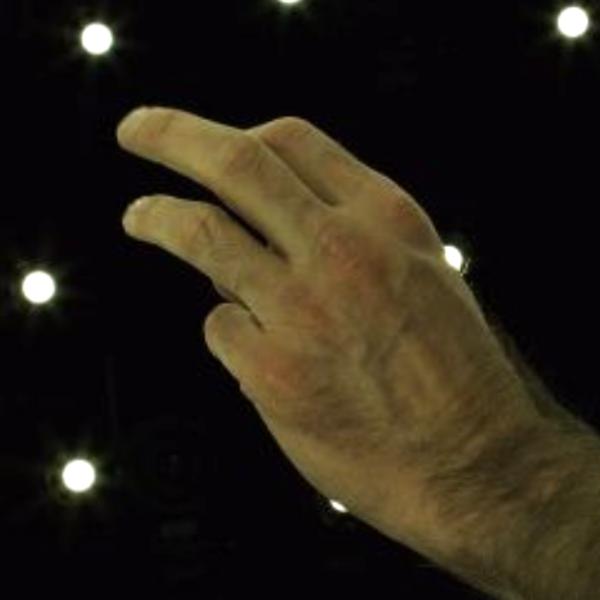}}
\hfill
\subfloat[\label{fig:temporal_orderingc}]{\includegraphics[width=0.24\linewidth]{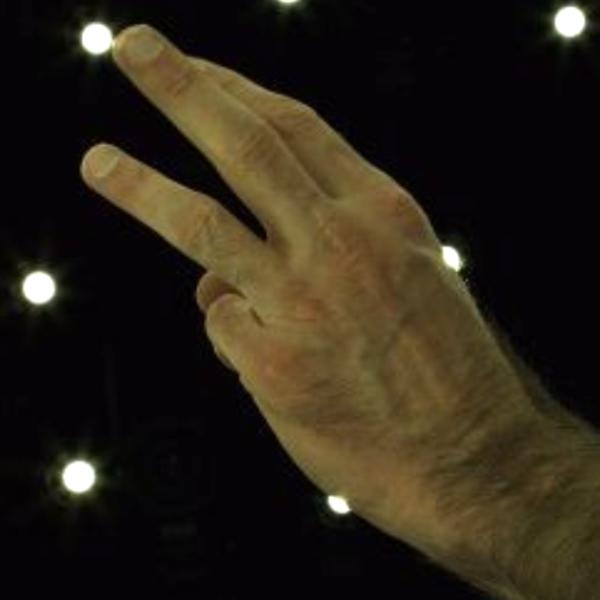}}
\hfill
\subfloat[\label{fig:temporal_orderingd}]{\includegraphics[width=0.24\linewidth]{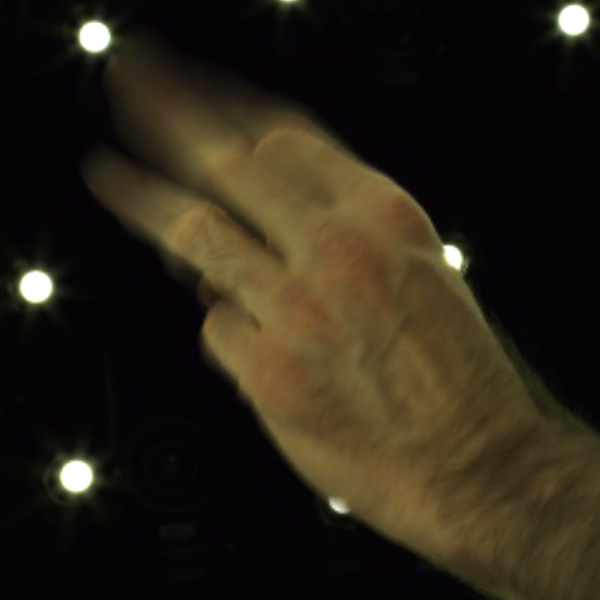}}
\end{center}
\vspace{-6mm}
\caption{\textbf{The ambiguity on temporal ordering.}
Hand image sequences of the extending~<(a)$\xrightarrow{}$(b)$\xrightarrow{}$(c)> and the folding <(c)$\xrightarrow{}$(b)$\xrightarrow{}$(a)> make the same result blur image~(d).
}
\vspace{-3mm}
\label{fig:temporal_ordering}
\end{figure}

Among the three sequential features, the hand feature at the middle of the motion ~$\mathbf{F}_{\text{M}}$ can be specified as similar to the conventional deblurring approaches~\cite{Nah_2017_CVPR, zamir2022restormer}.
However, we can not identify whether the hand at each end (\ie, $\mathbf{F}_{\text{E1}}$ or $\mathbf{F}_{\text{E2}}$) is come from the initial or final location of the motion due to the temporal ambiguity~\cite{Jin_2018_CVPR,purohit2019bringing, Zhang_2020_ACMMM, PAN_2019_CVPR}.
For example, suppose that we obtain the blurry hand image shown in Figure~\ref{fig:temporal_orderingd}.
Then we can not determine whether the blurry hand image comes from the motion of extending or folding.
In that regard, Unfolder outputs hand features from both ends of the motion (\ie, $\mathbf{F}_{\text{E1}}$ and $\mathbf{F}_{\text{E2}}$) without considering temporal order.
We note that exploiting the temporal information still benefits without explicitly considering the temporal order, and can further be stably optimized with the training loss introduced in Section~\ref{ssec:training_loss}.
%
%

\noindent\textbf{Extracting temporal joint features.}
From produced three sequential hand features, we extract the corresponding joint features, which contain essential hand articulation information~\cite{moon2022accurate} that helps to recover 3D hand meshes.
We first project the sequential hand features~$\mathbf{F}_{\text{E1}}$, $\mathbf{F}_{\text{E2}}$, and $\mathbf{F}_{\text{M}}$ into $dJ$ dimensional feature through $1 \times 1$ convolution layer, and reshape them into 3D heatmaps with the dimension of $\mathbb{R}^{J \times h \times w \times d}$, where $d=32$ is a depth discretization size and $J=21$ is a number of hand joints.
Then, we perform a soft-argmax operation~\cite{sun2018integral} on each heatmap to obtain the 3D joint coordinates of three temporal hands, $\mathbf{J_{\text{E1}}}$, $\mathbf{J_{\text{E2}}}$, and $\mathbf{J_{\text{M}}}$ with dimension of $\mathbb{R}^{J \times 3}$.
Using 3D joint coordinates in each temporal hand, we perform grid sampling~\cite{jaderberg2015spatial, moon2022accurate} on the corresponding feature map.
%
By doing so, we obtain temporal joint features $\mathbf{F}_{\text{J}_{\text{E1}}}$, $\mathbf{F}_{\text{J}_{\text{E2}}}$, and $\mathbf{F}_{\text{J}_{\text{M}}}$ with a dimension of $\mathbb{R}^{J \times c}$, which enable the following module to exploit temporal information effectively.

\begin{figure}[t]
\begin{center}
\includegraphics[width=0.98\linewidth]{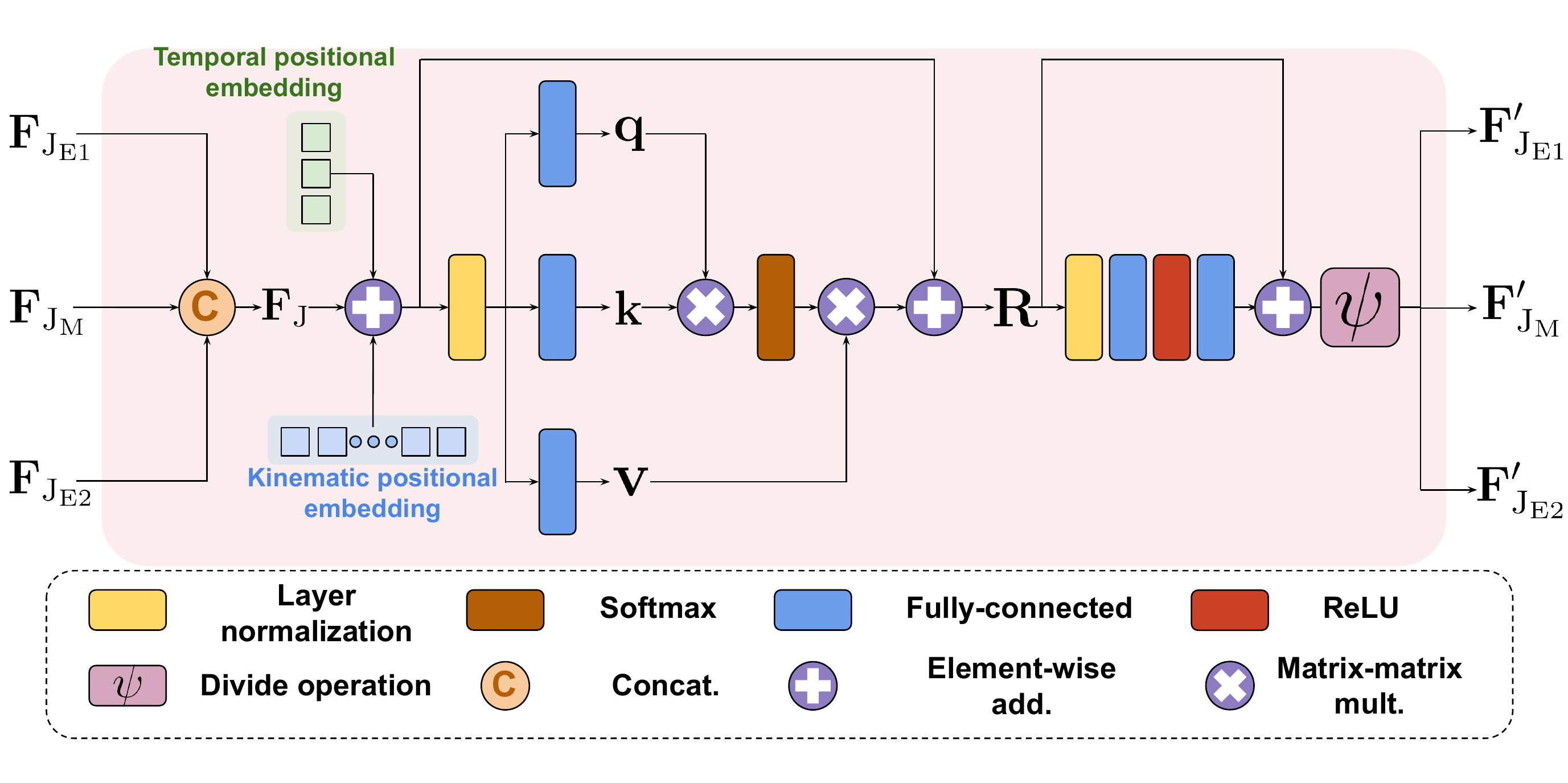}
\end{center}
\vspace{-8mm}
\caption{\textbf{Overall architecture of KTFormer.}
KTFormer refines temporal joint features $\mathbf{F}_{\text{J}_{\text{E1}}}$, $\mathbf{F}_{\text{J}_{\text{E2}}}$, and $\mathbf{F}_{\text{J}_{\text{M}}}$.
First, kinematic and temporal positional embeddings are introduced.
Then, the following self-attention mechanism refines joint features by leveraging attentive correlation between them, producing $\mathbf{F}^{\prime}_{\text{J}_{\text{E1}}}$, $\mathbf{F}^{\prime}_{\text{J}_{\text{E2}}}$, and $\mathbf{F}^{\prime}_{\text{J}_{\text{M}}}$.}
\vspace{-3mm}
\label{fig:tjformer}
\end{figure}
\subsection{KTFormer}
\label{ssec:ktformer}
\noindent\textbf{Kinematic-temporal positional embedding.}
The illustration of KTFormer is shown in Figure~\ref{fig:tjformer}.
KTFormer is the Transformer~\cite{vaswani2017attention}-based module that refines the joint feature~$\mathbf{F}_{\text{J}_{\text{E1}}}$, $\mathbf{F}_{\text{J}_{\text{E2}}}$, and $\mathbf{F}_{\text{J}_{\text{M}}}$ by considering the correlation between not only \emph{joints at the same time step} but also \emph{joints at different time steps}.
To utilize the temporal joint features as an input of Transformer, we first concatenate the three features along the joint dimension, producing $\mathbf{F}_{\text{J}} \in \mathbb{R}^{3J \times c}$.
Then, a learnable positional embedding, namely kinematic and temporal positional embeddings, is applied to $\mathbf{F}_{\text{J}}$.
The kinematic positional embedding~$\in \mathbb{R}^{J \times c}$ is applied along the joints dimension, while the temporal positional embedding~$\in \mathbb{R}^{3 \times c}$ is applied along the temporal dimension.
The kinematic and temporal positional embedding provide relative positions in kinematic and temporal space, respectively.

\noindent\textbf{Joint feature refinement with self-attention.}
KTFormer performs self-attention within $\mathbf{F}_{\text{J}}$ by extracting query~$\mathbf{q}$, key~$\mathbf{k}$, and value~$\mathbf{v}$ through three fully-connected layers.
Following the formulation of the standard Transformer~\cite{vaswani2017attention}, refined joint features for sequential hands~$\mathbf{F}^{\prime}_{\text{J}_{\text{E1}}}$, $\mathbf{F}^{\prime}_{\text{J}_{\text{E2}}}$, and $\mathbf{F}^{\prime}_{\text{J}_{\text{M}}}$ are formulated as follows:
\begin{equation}
\text{Att}(\mathbf{q},\mathbf{k},\mathbf{v}) = \text{softmax}(\frac{{\mathbf{q}}{\mathbf{k}}^{T}}{\sqrt{d_{\mathbf{k}}}})\mathbf{v},
\label{eq:1}
\end{equation}
\vspace{-2mm}
\begin{equation}
\mathbf{R} = \mathbf{F}_{\text{J}} + \text{Att}(\mathbf{q},\mathbf{k},\mathbf{v}),
\label{eq:2}
\end{equation}
\vspace{-2mm}
\begin{equation}
\mathbf{F}^{\prime}_{\text{J}_{\text{E1}}},\mathbf{F}^{\prime}_{\text{J}_{\text{E2}}},\mathbf{F}^{\prime}_{\text{J}_{\text{M}}} = \psi(\mathbf{F}_{\text{J}} + \mathrm{MLP}(\mathbf{R})),
\label{eq:3}
\end{equation}
where $d_{\mathbf{k}}=512$ is the feature dimension of the key~$\mathbf{k}$, and $\mathbf{R}$ is the residual feature.
$\mathrm{MLP}$ denotes multi-layer perceptron, and $\psi$  denotes a dividing operation, which separates features in dimension~$\mathbb{R}^{3J \times c}$ to three $\mathbb{R}^{J \times c}$.
Consequently, three joint features~$\mathbf{F}^{\prime}_{\text{J}_{\text{E1}}}$, $\mathbf{F}^{\prime}_{\text{J}_{\text{E2}}}$, and $\mathbf{F}^{\prime}_{\text{J}_{\text{M}}}$ are obtained by attentively utilizing kinematic and temporal information.

\subsection{Regressor}
The Regressor produces MANO~\cite{romero2022embodied} shape~(\ie, ${\beta}_{\text{E1}}$, ${\beta}_{\text{E2}}$, and ${\beta}_{\text{M}}$) and pose ~(\ie, ${\theta}_{\text{E1}}$, ${\theta}_{\text{E2}}$, and ${\theta}_{\text{M}}$) parameters, which correspond to sequential hands.
We describe the regression process of the middle hand (\ie, ${\beta}_{\text{M}}$ and ${\theta}_{\text{M}}$) as a representative procedure, and note that the process at different timesteps can be obtained in the same manner.
First, the shape parameter~${\beta}_{\text{M}}$ is obtained by forwarding the hand feature~$\mathbf{F}_{\text{M}}$ to a fully-connected layer after global average pooling~\cite{lin2013network}.
Second, the pose parameter~${\theta}_{\text{M}}$ is obtained by considering the kinematic correlation between hand joints.
To this end, we first concatenate refined joint feature~$\mathbf{F}^{\prime}_{\text{J}_{\text{M}}}$ with corresponding 3D coordinates~$\mathbf{J_{\text{M}}}$.
Then, we flatten the concatenated feature into one-dimensional vector~$\mathbf{f}_{\text{M}} \in \mathbb{R}^{J(c+3)}$.
Instead of regressing poses of entire joints from~$\mathbf{f}_{\text{M}}$ at once, the Regressor gradually estimates pose for each joint along the hierarchy of hand kinematic tree, following \cite{wan2021encoder}.
In detail, for a specific joint, its ancestral pose parameters and $\mathbf{f}_{\text{M}}$ are concatenated, and forwarded to a fully-connected layer to regress the pose parameters.
By adopting the same process for both ends, three MANO parameters are obtained from the Regressor.
Then, the MANO parameters are forwarded to the MANO layer to produce 3D hand meshes~$\mathbf{V}_{\text{E1}}$, $\mathbf{V}_{\text{E2}}$, and $\mathbf{V}_{\text{M}}$, where each denotes to meshes at both ends and middle, respectively.

\subsection{Training loss}
\label{ssec:training_loss}
During the training, a prediction on the middle of the motion can be simply supervised with GT of the middle frame.
On the other hand, it is ambiguous to supervise both ends of motion as the temporal order is not uniquely determined, as shown in Figure~\ref{fig:temporal_ordering}.
%
To resolve such temporal ambiguity during the loss calculation, we propose two items.
First, we employ \emph{temporal order-invariant loss}, which is invariant to GT temporal order~\cite{Rozumnyi2021defmo}.
To be specific, the temporal order in our loss function is determined in the direction that minimizes loss functions, not by the GT temporal order.
Second, we propose to use a \emph{Unfolder-driven temporal ordering}.
It determines the temporal order based on the output of Unfolder, then uses the determined temporal order to supervise the outputs of Regressor rather than determining the temporal order of two modules separately.
The effectiveness of the two items is demonstrated in the experimental section.

%
%
The overall loss function $\mathcal{L}$ is defined as follows:
\vspace*{-1mm}
\begin{equation}
\begin{split}
\mathcal{L} & = \mathcal{L}_\text{U} + \mathcal{L}_\text{R} \\
 & = \mathcal{L}_\text{U,M} + \mathcal{L}_\text{U,E} + \mathcal{L}_\text{R,M} + \mathcal{L}_\text{R,E},
\end{split}
\end{equation}
where $\mathcal{L}_\text{U}$ and $\mathcal{L}_\text{R}$ are loss functions applied to outputs of the Unfolder and the Regressor, respectively.
%
The subscripts M and E stand for prediction of the middle and both ends.

%
To supervise outputs of Unfolder, we define $\mathcal{L}_\text{U,M}$ and $\mathcal{L}_\text{U,E}$ as follows:
\begin{equation}
\mathcal{L}_\text{U,M} = \mathcal{L}_\text{joint}(\mathbf{J_{\text{M}}}, \mathbf{J^*_{\text{middle}}}),
\end{equation}
\vspace{-6mm}
\begin{alignat}{2}~\label{eq:loss_unfolder_end}
\mathcal{L}_\text{U,E} = \operatorname*{min} (~&\mathcal{L}_{\text{joint}}(\mathbf{J_{\text{E1}}}, \mathbf{J^*_{\text{initial}}}) &+~&~\mathcal{L}_{\text{joint}}(\mathbf{J_{\text{E2}}}, \mathbf{J^*_{\text{final}}}), \nonumber\\
&\mathcal{L}_{\text{joint}}(\mathbf{J_{\text{E1}}}, \mathbf{J^*_{\text{final}}}) &+ ~&~\mathcal{L}_{\text{joint}}(\mathbf{J_{\text{E2}}}, \mathbf{J^*_{\text{initial}}})~),
\end{alignat}
where $\mathbf{J^*_{\text{middle}}}$, $\mathbf{J^*_{\text{initial}}}$, and $\mathbf{J^*_{\text{final}}}$ are GT 3D joint coordinates of the middle, initial and final frame, respectively. $\mathcal{L}_{\text{joint}}$ is $L1$ distance between  predicted and GT joint coordinates.
%
The temporal order is determined to \emph{forward} if the first term in $\mathrm{min}$ of Eq.~\ref{eq:loss_unfolder_end} is selected as minimum and \emph{backward} otherwise.
Therefore, our loss function is \emph{invariant to the temporal order of GT}.
%

%
To supervise the outputs of the Regressor, we define the loss function $\mathcal{L}_\text{R,M}$ and $\mathcal{L}_\text{R,E}$ as follows: 
\begin{equation}
\mathcal{L}_\text{R,M} = \mathcal{L}_{\text{mesh}}(\Theta_\text{M}, \Theta^*_\text{middle}),
\end{equation}
\vspace{-6mm}
\begin{alignat}{5}
        \mathcal{L}_\text{R,E}=
          &\mathds{1}_\textbf{f} &~( &~\mathcal{L}_{\text{mesh}}(\Theta_\text{E1}, \Theta^*_\text{initial})  & +~ &\mathcal{L}_{\text{mesh}}(\Theta_\text{E2}, \Theta^*_\text{final}) &~) \nonumber \\
        + &\mathds{1}_\textbf{b} &~( &~\mathcal{L}_{\text{mesh}}(\Theta_\text{E1}, \Theta^*_\text{final})&+~ &\mathcal{L}_{\text{mesh}}(\Theta_\text{E2}, \Theta^*_\text{initial})   &~),
    \label{eq:loss_with_order}
\end{alignat}
where $\mathds{1}_\textbf{f} = 1$ when the temporal order is determined to forward in Eq.~\ref{eq:loss_unfolder_end} otherwise 0, and $\mathds{1}_\textbf{b} = 1 - \mathds{1}_\textbf{f}$. 
In other words, the temporal order of Eq.~\ref{eq:loss_with_order} follows that of Eq.~\ref{eq:loss_unfolder_end}, which we call \emph{Unfolder-driven temporal ordering}.
$\Theta_\bullet = \{\theta_\bullet, \beta_\bullet\}$ is GT or predicted MANO parameters, where the superscript $^*$ denotes GT.
$\mathcal{L}_{\text{mesh}}$ is the summation of three $L1$ distances between GT and prediction of: 1) MANO parameters 2) 3D joint coordinates obtained by multiplying joint regression matrix to hand mesh 3) 2D joint coordinates projected from 3D joint coordinates. 

\section{Experiments}

\subsection{Datasets and evaluation metrics}

\noindent\textbf{BlurHand.}
The BlurHand~(BH) is our newly presented 3D hand pose dataset containing realistic blurry hand images as introduced in Section~\ref{sec:blurhand_dataset}.
%
We train and test the 3D hand mesh estimation networks on the train and test splits of the BH.
\begin{table}
\small
\centering
\setlength\tabcolsep{1.0pt}
\def\arraystretch{1.1}
\newcommand{\spacing}{\;}
\resizebox{0.8\linewidth}{!}{
\begin{tabular}{L{3.2cm}|C{1.5cm}|C{1.5cm}|C{1.5cm}}
\specialrule{.1em}{.05em}{.05em}
\hline
\,~\multirow{2}{*}{Methods} & \multirow{2}{*}{Train set} & \multicolumn{2}{c}{Test set} \\
\cline{3-4}
& & IH2.6M & BH \\\hline
\spacing~\multirow{2}{*}{I2L-MeshNet~\cite{moon2020i2l}} & IH2.6M & 22.27 & 29.16 \\
& BH & 24.30 & 24.32  \\ \hline
\spacing~\multirow{2}{*}{METRO~\cite{lin2021end}} & IH2.6M  & 18.44 & 35.43 \\
& BH & 20.19 & 20.54   \\ \hline
\spacing~\multirow{2}{*}{Pose2Pose~\cite{moon2022accurate}} & IH2.6M  & 16.85 & 25.36 \\
& BH & 18.40 & 18.80 \\ \hline
\spacing~\multirow{2}{*}{\textbf{BlurHandNet (Ours)}} & IH2.6M  & \textbf{15.33} & 24.57 \\
& BH & 16.12 & \textbf{16.80} \\ \hline
\specialrule{.1em}{.05em}{.05em}
\end{tabular}}
\vspace{-3mm}
\caption{
    \textbf{Effectiveness of BlurHand on handling blurry hand.}
    We calculate MPJPE~(mm) on hand meshes located in the middle of the motion.
    %
    %
}
\vspace{-3mm}
\label{table:baseline_performance}
\end{table}

\begin{figure}[t]
\begin{center}
\includegraphics[width=1.\linewidth]{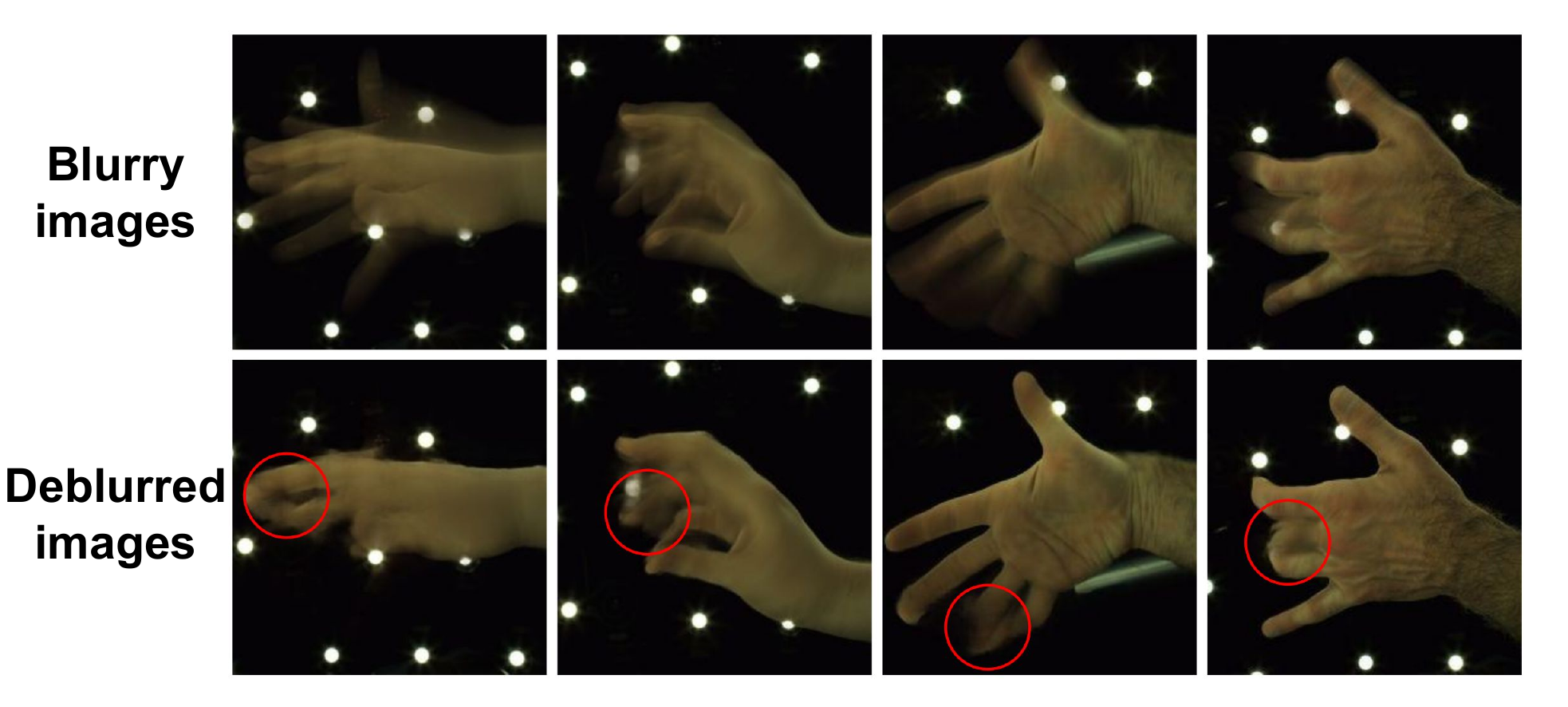}
\end{center}
\vspace{-8mm}
\caption{\textbf{Examples of deblurred images.} Since the blurry hand undergoes challenging blur from complex articulation, even the state-of-the-art deblurring method~\cite{chen2022simple} cannot fully restore and blur artifact remains, highlighted with red circles.
}
\vspace{-2mm}
\label{fig:deblur_example}
\end{figure}

\noindent\textbf{InterHand2.6M.}
InterHand2.6M~\cite{moon2020interhand2} (IH2.6M) is a recently presented large-scale 3D hand dataset.
It is captured under highly calibrated camera settings and provides accurate 3D annotations for hand images.
We employ IH2.6M as a representative of sharp hand frames, as the hand images in IH2.6M do not contain blur.
We use a subset of IH2.6M for training and testing purposes, where the subset is a set of the third sharp frame in \figref{manufacture_diagram} for each image of BH.

%
%
%
%
%
%


\noindent\textbf{YT-3D.}
YouTube-3D-hands~(YT-3D)~\cite{kulon2020weakly} is a 3D hand dataset with diverse and non-laboratory videos collected from youtube.
We utilize the YT-3D as an additional training dataset when testing on YT-3D.
Since YT-3D does not provide 3D GTs, we only provide a qualitative comparison of this dataset without quantitative evaluations.

\noindent\textbf{Evaluation metrics.}
We use mean per joint position error (MPJPE) and mean per vertex position error (MPVPE) as our evaluation metrics.
The metrics measure Euclidean distance (mm) between estimated coordinates and groundtruth coordinates.
Before calculating the metrics, we align the translation of the root joint (\ie, wrist).

\subsection{Ablation study}
\begin{table}
\small
\centering
\setlength\tabcolsep{1.0pt}
\def\arraystretch{1.1}
\newcommand{\spacing}{\;}
\resizebox{0.90\linewidth}{!}{
\begin{tabular}{L{3.2cm}|C{1.8cm}|C{1.8cm}|C{1.5cm}|C{1.5cm}}
\specialrule{.1em}{.05em}{.05em}
\hline
\,~\multirow{2}{*}{Methods} & \multirow{2}{*}{Train set} & \multirow{2}{*}{Test set} & MPJPE & MPVPE \\
& & & (mm) & (mm) \\
\hline
\spacing~\multirow{3}{*}{I2L-MeshNet~\cite{moon2020i2l}} & IH2.6M & BH+Deblur & 26.56 & 25.23 \\
& BH+Deblur & BH+Deblur & 26.13 & 25.00 \\
& BH & BH & 24.32 & 23.08 \\
\hline
\spacing~\multirow{3}{*}{METRO~\cite{lin2021end}} & IH2.6M & BH+Deblur & 26.07 & 32.05 \\
& BH+Deblur & BH+Deblur & 20.11 & 26.55 \\
& BH & BH & 20.54 & 27.03 \\
\hline
\spacing~\multirow{3}{*}{Pose2Pose~\cite{moon2022accurate}} & IH2.6M & BH+Deblur & 22.43 & 21.04 \\
& BH+Deblur & BH+Deblur & 18.81 & 17.43 \\
& BH & BH & 18.80 & 17.42 \\
\hline
\spacing~\multirow{3}{*}{\textbf{BlurHandNet (Ours)}} & IH2.6M& BH+Deblur & 21.37 & 19.93 \\
& BH+Deblur & BH+Deblur & 17.28 & 15.82  \\
& BH & BH & \textbf{16.80} & \textbf{15.30} \\
\hline
\specialrule{.1em}{.05em}{.05em}
\end{tabular}}
\vspace{-3mm}
\caption{
    \textbf{Effectiveness of BlurHand compared to deblurring baseline.}
    MPJPE and MPVPE are calculated at hand meshes located in the middle of the motion.
}
\vspace{-2mm}
\label{table:comparison_sota}
\end{table}

\begin{figure}
    \centering
    \newcommand{\spacing}{0.195}
    \captionsetup[subfigure]{labelfont=scriptsize, textfont=scriptsize}
        \subfloat{\includegraphics[width=\spacing\linewidth]{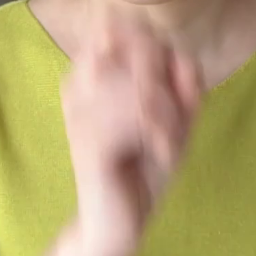}}
        \hfill
        \subfloat{\includegraphics[width=\spacing\linewidth]{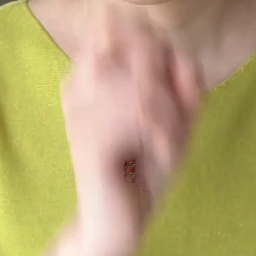}}
        \hfill
        \subfloat{\includegraphics[width=\spacing\linewidth]{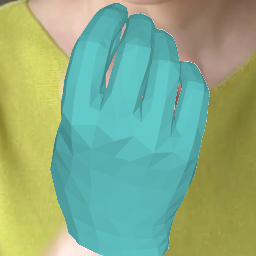}}
        \hfill
        \subfloat{\includegraphics[width=\spacing4\linewidth]{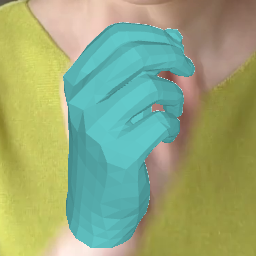}}
        \hfill
        \subfloat{\includegraphics[width=\spacing4\linewidth]{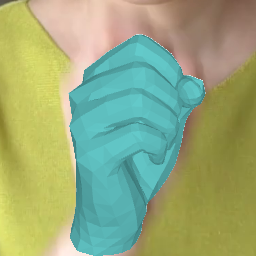}}
    \addtocounter{subfigure}{-5}
    \\
        \subfloat[Blurry]{\includegraphics[width=\spacing\linewidth]{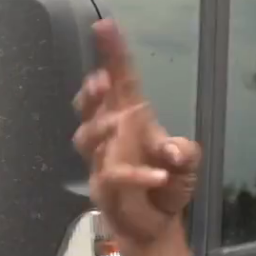}}
        \hfill
        \subfloat[Deblurred\label{fig:comparision_yt3db}]{\includegraphics[width=\spacing\linewidth]{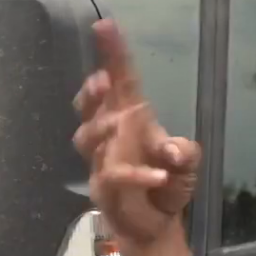}}
        \hfill
        \subfloat[IH2.6M\label{fig:comparision_yt3dc}]{\includegraphics[width=\spacing\linewidth]{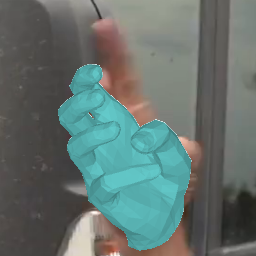}}
        \hfill
        \subfloat[BH+\textbf{D}\label{fig:comparision_yt3dd}]{\includegraphics[width=\spacing\linewidth]{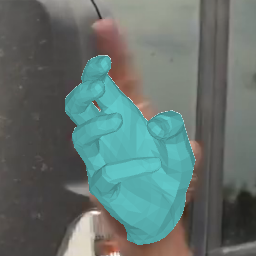}}
        \hfill
        \subfloat[BH+YT3D\label{fig:comparision_yt3de}]{\includegraphics[width=\spacing\linewidth]{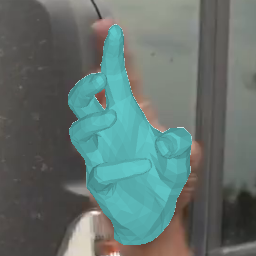}}
    \\
\vspace{-1mm}
\captionsetup[subfigure]{labelformat=empty}
\begin{minipage}{1.0\linewidth}
\subcaption{\textcolor{white}{AAAA}image\textcolor{white}{AAAAAA}image\textcolor{white}{AAAAA}+YT3D\textcolor{white}{AAAAA}+YT3D\textcolor{white}{AAAAASAAAAAA}}
\end{minipage}
\\
\vspace{-2mm}
    %
    \caption{\textbf{Qualitative comparison on real-world blurry hand images.}
    %
    The captions below figures describe datasets used to train 3D hand mesh estimation networks.
    %
    A setting with \textbf{D} represents that the network is trained on deblurred BH and tested on (b).
    %
    }
    \vspace{-0.4cm}
    \label{fig:comparison_yt3d}
\end{figure}

\noindent\textbf{Benefit of BlurHand dataset.}
%
Directly measuring how much synthesized blur is close to the real one is still an open research problem in the deblurring community~\cite{zhang2020deblurring,rim_2022_ECCV}. 
Hence, we justify the usefulness of the presented BlurHand using indirect commonly used protocols~\cite{rim_2022_ECCV}, \ie, train the model with the presented dataset and test it on unseen blurry images.
%
%
Table~\ref{table:baseline_performance} shows that all 3D hand mesh recovery networks trained on InterHand2.6M suffer severe performance drops when they are tested on BlurHand, while networks trained on BlurHand perform well.
These experimental results validate that training on BlurHand is necessary when handling the blurry hand.
In addition, networks trained on BlurHand also perform well on InterHand2.6M, which consists of sharp images.
This shows the generalizability of our dataset to sharp images.
%

Table~\ref{table:comparison_sota} shows that utilizing presented BlurHand is more valuable than applying deblurring methods~\cite{chen2022simple}.
The deblurring method~\cite{chen2022simple}, trained on our BlurHand as a pre-trained deblurring network, performs poorly on our BlurHand.
Moreover, the networks trained on sharp images~(IH2.6M) or deblurred images and tested on deblurred images perform worse than those trained and tested on our BlurHand.
Such comparisons demonstrate the usefulness of our BlurHand dataset compared to applying deblurring methods.
One reason is that, as Figure~\ref{fig:deblur_example} shows, deblurring methods often fail to restore sharp hand images due to complicated hand motions.
Another reason is that deblurring removes temporal information from the blurry image, which is helpful for reconstructing accurate 3D hand mesh sequences.

Figure~\ref{fig:comparison_yt3d} provides a qualitative comparison of real-world blurry images in YT-3D, which further demonstrates the usefulness of our BlurHand.
We train BlurHandNet on three different combinations of datasets: 1) InterHand2.6M and YT-3D (\ref{fig:comparision_yt3db}), 2) deblurred BlurHand and YT-3D (\ref{fig:comparision_yt3dc}), and 3) BlurHand and YT-3D (\ref{fig:comparision_yt3dd}).
The comparison shows that networks trained on BlurHand produce the most robust 3D meshes, demonstrating the generalizability of BlurHand.

\begin{table}
\small
\centering
\setlength\tabcolsep{1.0pt}
\def\arraystretch{1.1}
\resizebox{0.8\linewidth}{!}{
\begin{tabular}{C{1.4cm}|C{1.4cm}|C{1.4cm}|C{1.1cm}|C{1.1cm}|C{1.1cm}}
\specialrule{.1em}{.05em}{.05em}
\hline
\multirow{2}{*}{Deblur} & \multirow{2}{*}{Unfolder} & \multirow{2}{*}{KTFormer} & \multicolumn{3}{c}{MPJPE} \\
\cline{4-6}
& &  & initial & middle & final \\
\hline
\xmark & \xmark & \xmark & - & 17.89 & - \\
\xmark & \xmark & \cmark & - & 17.41 & - \\
\xmark & \cmark & \xmark & 18.94 & 17.55 & 19.05\\
\xmark & \cmark & \cmark & \textbf{18.08} & \textbf{16.80} & \textbf{18.21} \\\hline
\cmark & \xmark & \xmark & - & 17.28 & - \\ 
\cmark & \cmark & \cmark & 18.95 & 17.28 & 19.10 \\
\hline
\specialrule{.1em}{.05em}{.05em}
\end{tabular}}
\vspace{-3mm}
\caption{
    \textbf{Ablation study on proposed Unfolder and KTFormer.}
    %
    (\cmark~in Deblur): The experiments with the Deblur item checked are trained and tested on deblurred BlurHand.
    (Second row): We only employ features from a single time step when applying KTFormer, as temporal information does not exist without Unfolder.
}
\vspace{-3mm}
\label{table:ablation_each_modules}
\end{table}


\begin{table}[!t]
\small
\centering
\setlength\tabcolsep{1.0pt}
\def\arraystretch{1.1}
\resizebox{0.85\linewidth}{!}{
\begin{tabular}{L{2.6cm}|C{1.1cm}|C{1.1cm}|C{1.1cm}|C{1.1cm}|C{1.1cm}}
\specialrule{.1em}{.05em}{.05em}
\hline
\,~\multirow{2}{*}{\# of unfolding} &  \multicolumn{5}{c}{MPJPE} \\
\cline{2-6}
& initial & initial$^*$ & middle & final$^*$ & final \\
\hline
\;~1 & - & - & 17.41 & - & - \\
\;~\textbf{3~(BlurHandNet)} & 18.08 & - & 16.80 & - & 18.21 \\
\;~5 & 18.06 & 17.18 & 16.78 & 17.36 & 18.18 \\
\hline
\specialrule{.1em}{.05em}{.05em}
\end{tabular}}
\vspace{-3mm}
\caption{
    \textbf{Ablation study on the number of unfolded hands.}
    The initial$^*$ and final$^*$ denote hands between the initial and middle, and the middle and final, respectively.
}
\vspace{-3mm}
\label{table:unfoldingnumber}
\end{table}

\noindent\textbf{Effectiveness of Unfolder and KTFormer.}
Table~\ref{table:ablation_each_modules} shows that using both Unfolder and KTFormer improves 3D mesh estimation accuracy by a large margin.
As the proposed Unfolder allows a single image to be regarded as three sequential hands, we evaluate hands in both ends and middle of the motion.
Since the temporal order of hand meshes in both ends~(\ie, $\mathbf{V}_{\text{E1}}$ and $\mathbf{V}_{\text{E2}}$) is not determined, we report better MPJPE among the initial-final and final-initial pairs following~\cite{Rozumnyi2021defmo, Argaw_2021_CVPRW}.
Solely employing one of Unfolder or KTFormer~(the second and third rows) shows a slight improvement over the baseline network, which is designed without any of the proposed modules~(the first row).
On the other hand, our BlurHandNet~(the fourth row) results in great performance boosts, by benefiting from the combination of two modules that effectively complement each other.
In particular, KTFormer benefits from temporal information which is provided by Unfolder.
Consequently, introducing both Unfolder and KTFormer, which have strong synergy, consistently improves the 3D errors in all time steps.

In the point of the baseline~(the first row), using our two proposed modules, Unfolder and KTFormer, leads to more performance gain~(the fourth row) than training and testing a network on deblurred BlurHand~(the fifth row).
This comparison shows that utilizing proposed modules is more effective than using deblurring methods.
Interestingly, using our two modules does not bring performance gain when training and testing on deblurred BlurHand~(the last row) compared to the deblur baseline~(the fifth row).
This validates our statement that deblurring prohibits networks from utilizing temporal information.

\begin{figure*}[t]
\begin{center}
\includegraphics[width=0.81\linewidth]{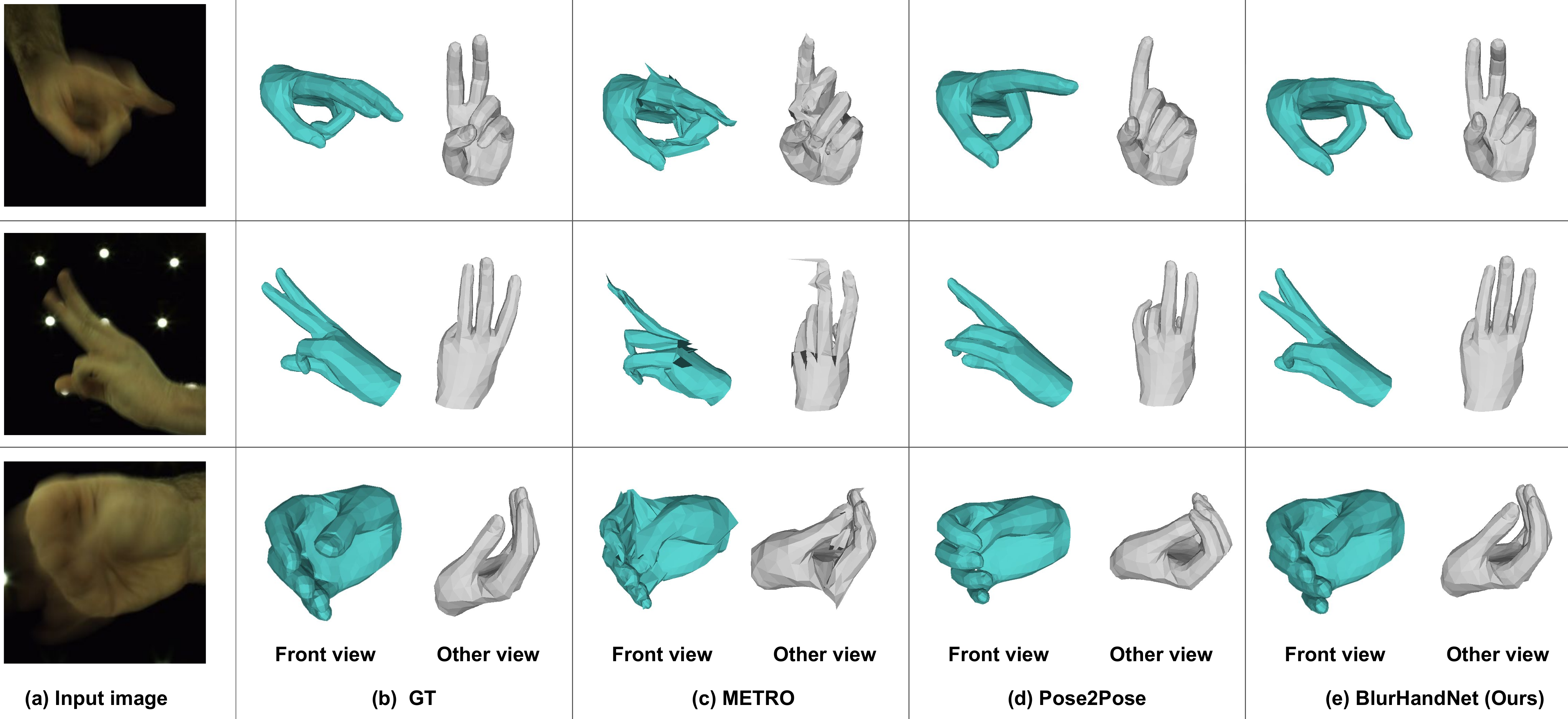}
\end{center}
\vspace*{-5mm}
\caption{\textbf{Visual comparison of the BlurHandNet and state-of-the-art 3D hand mesh estimation methods~\cite{lin2021end,moon2022accurate} on BlurHand.}}
\vspace*{-6mm}
\label{fig:comparison_interhand2}
\end{figure*}
\begin{figure}[t]
\begin{center}
\subfloat[Input image \label{fig:attn_visa}]{\includegraphics[width=0.37\linewidth]{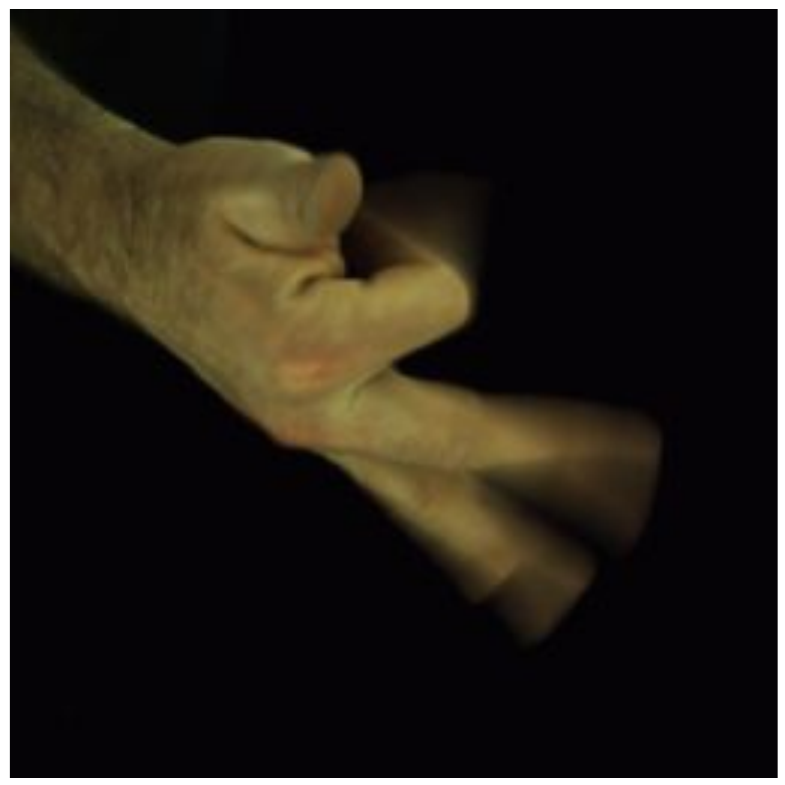}}
\hspace{2mm}
\subfloat[Attention map \label{fig:attn_visb}]{\includegraphics[width=0.52\linewidth]{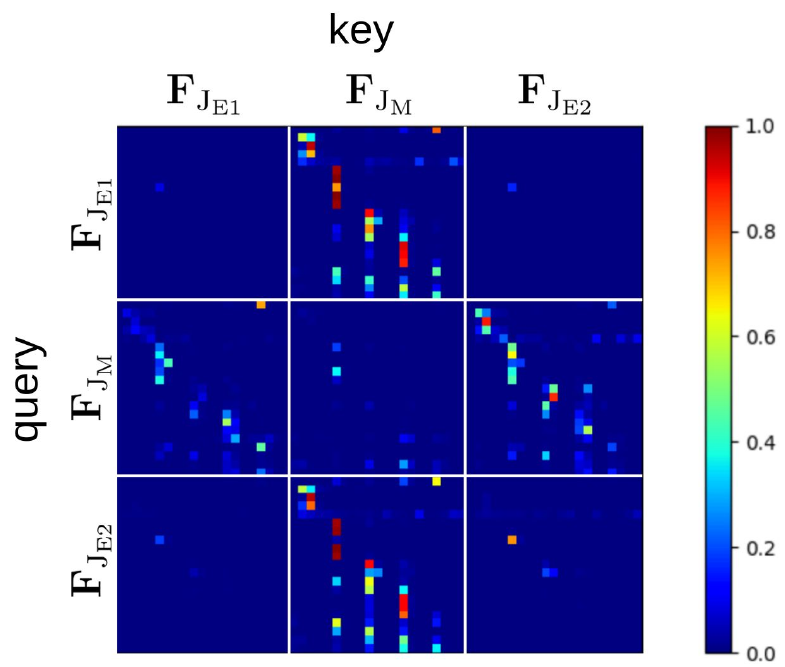}}
\end{center}
\vspace{-6mm}
\caption{\textbf{Visualization of attention map from KTFormer.}
The pixel at location~($i$,$j$) represents the correlation between \emph{i-th} joint feature and \emph{j-th} joint feature.
The attention map is obtained after performing a softmax operation across each row (query).
}
\vspace{-5mm}
\label{fig:attn_vis}
\end{figure}

\noindent\textbf{Effect of the number of unfolded hands.}
Table~\ref{table:unfoldingnumber} shows that unfolding more sequential hands further improves the 3D errors.
As our KTFormer utilizes temporal information to enhance the joint feature, the number of hand sequences can affect the overall performance.
Although unfolding a blurry hand into five sequential hands shows the best results, the performance is nearly saturated when a blurry hand is unfolded into three sequential hands.
Considering the increased computational costs of producing additional hands and the temporal input size of KTFormer, we design our Unfolder to produce three sequential hands.
We note that our BlurHandNet can be easily extended if more number of unfolding is needed for some applications.


\begin{table}
\small
\centering
\setlength\tabcolsep{1.0pt}
\def\arraystretch{1.1}
\resizebox{0.65\linewidth}{!}{
\begin{tabular}{C{1.4cm}|C{1.4cm}|C{1.1cm}|C{1.1cm}|C{1.1cm}}
\specialrule{.1em}{.05em}{.05em}
\hline
\multirow{2}{*}{Kinematic} & \multirow{2}{*}{Temporal} & \multicolumn{3}{c}{MPJPE} \\
\cline{3-5}
& & initial & middle & final \\
\hline
\xmark & \xmark & 18.99 & 17.79 & 19.06 \\
\cmark & \xmark & 18.28 & 16.92 & 18.34 \\
\xmark & \cmark & 18.79 & 17.41 & 18.92\\
\cmark & \cmark & \textbf{18.08} & \textbf{16.80} & \textbf{18.21} \\
\hline
\specialrule{.1em}{.05em}{.05em}
\end{tabular}}
\vspace{-3mm}
\caption{\textbf{Ablation study on the kinematic and temporal positional embeddings.} 
}
\vspace{-3mm}
\label{table:positional_embedding}
\end{table}
\noindent\textbf{Effect of the kinematic and temporal positional embeddings.}
Table~\ref{table:positional_embedding} shows that our positional embedding setting, which uses both kinematic and temporal positional embedding, achieves the best performance.
We design four variants with different positional embedding settings.
The second and third rows, where either one of kinematic and temporal positional embedding is applied, achieve better results than a baseline without any positional embedding (the first row), but worse results than ours (the last row).
This indicates that positional information of both kinematic and temporal dimensions is necessary for KTFormer.

\begin{table}
\small
\centering
\setlength\tabcolsep{1.0pt}
\def\arraystretch{1.1}
\resizebox{0.8\linewidth}{!}{
\begin{tabular}{C{2.2cm}|C{2.4cm}|C{1.0cm}|C{1.0cm}|C{1.0cm}}
\specialrule{.1em}{.05em}{.05em}
\hline
Temporal order & Unfolder-driven & \multicolumn{3}{c}{MPJPE} \\
\cline{3-5}
invariant loss & temporal ordering & initial & middle & final \\
\hline
\xmark & \xmark & 18.72 & 16.98 & 18.86 \\
\cmark & \xmark & 18.44 & 17.14 & 18.55\\
\cmark & \cmark & \textbf{18.08} & \textbf{16.80} & \textbf{18.21} \\
\hline
\specialrule{.1em}{.05em}{.05em}
\end{tabular}}
\vspace{-3mm}
\caption{
    \textbf{Ablation study on proposed loss functions.} 
}
\vspace{-3mm}
\label{table:minimum}
\end{table}
\noindent\textbf{Effectiveness of the proposed loss functions.}
Table~\ref{table:minimum} shows that two proposed items in our loss function, \emph{temporal order-invariant loss} and \emph{Unfolder-driven temporal ordering} as introduced in Section~\ref{ssec:training_loss}, are necessary for the high performance.
We compare three variants for loss design at both ends of the motion, while keeping the loss function for the middle of the motion the same.
In detail, the settings without \emph{temporal order-invariant loss} are supervised with 3D meshes following the GT temporal order instead of determining the order based on Eq.~\ref{eq:loss_unfolder_end}.
On the other hand, the setting without \emph{Unfolder-driven temporal ordering} supervises Regressor with Regressor-driven temporal ordering, which indicates that Unfolder and Regressor can be supervised with different temporal ordering.
Since inferring the temporal order of GT is ambiguous, the setting without \emph{temporal order-invariant loss} degrades the performance as it forces to strictly follows the temporal order of GT.
Utilizing the proposed \textit{Unfolder-driven temporal ordering} performs the best, as it provides consistent temporal order to both Unfolder and Regressor, making the training stable.



\noindent\textbf{Visualization of attention map from KTFormer.}
Figure~\ref{fig:attn_vis} shows a visualized attention map, obtained by the self-attention operation in KTFormer.
Here, query and key are obtained from a combination of temporal joint features $\mathbf{F}_{\text{J}_{\text{E1}}}$, $\mathbf{F}_{\text{J}_{\text{C}}}$, and $\mathbf{F}_{\text{J}_{\text{E1}}}$, as described in Section~\ref{ssec:ktformer}.
The figure shows that our attention map produces two diagonal lines, representing a strong correlation between the corresponding query and key.
Specifically, features from both ends of motion,~$\mathbf{F}_{\text{J}_{\text{E1}}}$ and $\mathbf{F}_{\text{J}_{\text{E2}}}$~(the first and third queries), show high correlation with the middle hand feature~$\mathbf{F}_{\text{J}_{\text{M}}}$ (the second key), and $\mathbf{F}_{\text{J}_{\text{M}}}$~(the second query) shows high correlation with $\mathbf{F}_{\text{J}_{\text{E1}}}$ and $\mathbf{F}_{\text{J}_{\text{E2}}}$ (the first and third keys).
This indicates that temporal information is highly preferred to compensate for insufficient joint information in a certain time step.
This is also consistent with the result in Table~\ref{table:ablation_each_modules}.
In the second row of Table~\ref{table:ablation_each_modules}, solely employing KTFormer without Unfolder shows slight performance improvement over the baseline due to the lack of opportunity to exploit temporal information from both ends.

\subsection{Comparison with state-of-the-art methods}
Table~\ref{table:baseline_performance} and ~\ref{table:comparison_sota} show that our BlurHandNet outperforms the previous state-of-the-art 3D hand mesh estimation methods in all settings.
As the previous works~\cite{moon2020i2l,lin2021end,moon2022accurate} do not have a special module to address blurs, they fail to produce accurate 3D meshes from blurry hand images.
On the contrary, by effectively handling the blur using temporal information, our BlurHandNet robustly estimates the 3D hand mesh, even under abrupt motion.
Figure~\ref{fig:comparison_interhand2} further shows that our BlurHandNet produces much better results than previous methods on BlurHand.

\section{Conclusion}
We present the BlurHand dataset, containing realistic blurry hand images with 3D GTs, and the baseline network BlurHandNet.
Our BlurHandNet regards a single blurry hand image as sequential hands to utilize the temporal information from sequential hands, which makes the network robust to the blurriness.
Experimental results show that BlurHandNet achieves state-of-the-art performance in estimating 3D meshes from the newly proposed BlurHand and real-world test sets.

\noindent\textbf{Acknowledgments.}
This work was supported in part by the IITP grants [No.2021-0-01343, Artificial Intelligence Graduate School Program (Seoul National University), No.2022-0-00156, No. 2021-0-02068, and No.2022-0-00156], and the NRF grant [No. 2021M3A9E4080782] funded by the Korea government (MSIT).
%

{\small
\bibliographystyle{ieee_fullname}
\bibliography{egbib}
}

\end{document}



\long\def\ignorethis#1{}



\newcommand{\img}[1]{\mathbf{I}_{\text{#1}}}
\newcommand{\paren}[1]{\left( #1 \right)}
\newcommand{\bparen}[1]{\left[ #1 \right]}
\newcommand{\feature}[1]{\phi \paren{#1}}
\newcommand{\normtwo}[1]{\lVert #1 \rVert_2^2}
\newcommand{\normone}[1]{\left\lVert #1 \right\rVert_1}

\newcommand{\Paragraph}[1]{\vspace{1mm}\noindent\textbf{#1}}

\newcommand{\figref}[1]{Figure~\ref{fig:#1}}
\newcommand{\tabref}[1]{Table~\ref{tab:#1}} 
\newcommand{\itmref}[1]{[\ref{itm:#1}]}     
\newcommand{\eqnref}[1]{\eqref{eq:#1}}
\newcommand{\secref}[1]{Section~\ref{sec:#1}}
\newcommand{\eqmain}[1]{(\textcolor{blue}{#1})}
\newcommand{\fakeref}[1]{\textcolor{MyGreen}{#1}}
\newcommand{\fakeeqref}[1]{\textcolor{MyGreen}{(#1)}}

\newcommand{\mb}[1]{\mathbf{#1}}
\newcommand{\bs}[1]{\boldsymbol{#1}}
\newcommand{\n}{\mbox{\qquad}}              
\newcommand{\red}[1]{{\color{red}#1}}

\newcommand{\ignore}[1]{}   
\newcommand{\cmt}[1]{\begin{sloppypar}\large\textcolor{red}{#1}\end{sloppypar}}

\newcommand{\TODO}[1]{\textcolor{red}{[TODO]\{#1\}}}
\newcommand{\todo}[1]{\textcolor{red}{#1}}
\newcommand{\torevise}[1]{\textcolor{blue}{#1}}
\newcommand{\revise}[1]{\textcolor{blue}{#1}}
\newcommand{\copied}[1]{\textcolor{red}{[COPIED: #1]}}

\newcommand{\needref}{[\textcolor{blue}{put}, \textcolor{blue}{some}, \textcolor{blue}{references}]}

\newcommand{\tabspace}{\vspace{-2mm}}
\newcommand{\tabxspace}{\vspace{-4mm}}
\newcommand{\figspace}{\vspace{-2mm}}
\newcommand{\figxspace}{\vspace{-3mm}}
\newcommand{\best}[1]{{\textcolor{red}{\textbf{#1}}}}
\newcommand{\secondbest}[1]{{\textcolor{blue}{\underline{#1}}}}

\newcommand{\ts}{\textsuperscript}

\newcommand{\set}[1]{\{#1\}}


\def\PE{\Phi}

\newcommand{\mpage}[2]
{
\begin{minipage}{#1\linewidth}\centering
#2
\end{minipage}
}

\newcommand{\replace}[2]{\textcolor{red}{\sout{#1}} \textcolor{blue}{#2}}
\newcommand{\fix}[1]{\textcolor{red}{#1}}
\newcommand{\proposed}{UDMM}

\newcommand{\bsd}{MSD}

\newcommand{\blurhigh}{\mathbf{I}_{\hat{\text{B}}}}
\newcommand{\blurlow}{\mathbf{I}_{\hat{\text{B}},\hat{\text{LR}}}}

\definecolor{MyGreen}{cmyk}{100, 0, 100, 0}

\newcommand{\citenumber}[1]{[\textcolor{MyGreen}{#1}]}
\newcommand{\refnumber}[1]{\textcolor{red}{#1}}

\newcolumntype{L}[1]{>{\raggedright\let\newline\\\arraybackslash\hspace{0pt}}m{#1}}
\newcolumntype{C}[1]{>{\centering\let\newline\\\arraybackslash\hspace{0pt}}m{#1}}
\newcolumntype{R}[1]{>{\raggedleft\let\newline\\\arraybackslash\hspace{0pt}}m{#1}}

\newcommand{\jaeha}[1]{{\textcolor{BurntOrange}{\textbf{Jaeha: }#1}}}
\newcommand{\joonkyu}[1]{{\textcolor{blue}{\textbf{Joonkyu: }#1}}}
\newcommand{\younguk}[1]{{\textcolor{ProcessBlue}{\textbf{Younguk: }#1}}}


\title{\emph{Supplementary Material for} \\ Recovering 3D Hand Mesh Sequence from a Single Blurry Image: \\ A New Dataset and Temporal Unfolding}
\maketitle
\thispagestyle{empty}

\setcounter{section}{0}
\setcounter{figure}{0}
\setcounter{table}{0}
\setcounter{equation}{0}

\renewcommand{\thetable}{S\arabic{table}}
\renewcommand{\thesection}{S\arabic{section}}
\renewcommand{\thefigure}{S\arabic{figure}}
\renewcommand{\theequation}{S\arabic{equation}}

In this supplementary material, we provide 
more various visual results, discussions, and other details that could not be included in the main manuscript due to the lack of space.
The contents are summarized below:

\begin{compactitem}[$\bullet$]
    %
    \item \ref{sec:detail_blurhand}. Statistics on the BlurHand dataset
    \item \ref{sec_supp:deblurring}. Results from various deblurring methods
    \item \ref{sec:training_details}. Training details
    %
    \item \ref{sec:more_results}. Additional qualitative results
    \item \ref{sec:discussion}. Discussions
\end{compactitem}

\section{Statistics on the BlurHand dataset}
\label{sec:detail_blurhand}

\begin{figure}[t]
    \newcommand{\ww}{0.50}
    \centering
    \captionsetup[subfigure]{labelfont=scriptsize, textfont=scriptsize}
    \subfloat[Train set \label{motion_blur_a}]{\includegraphics[width=\ww \linewidth]{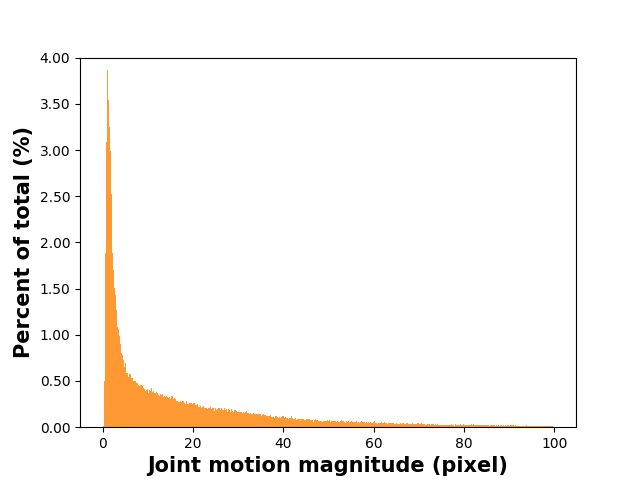}}
    \subfloat[Test set \label{motion_blur_b}]{\includegraphics[width=\ww \linewidth]{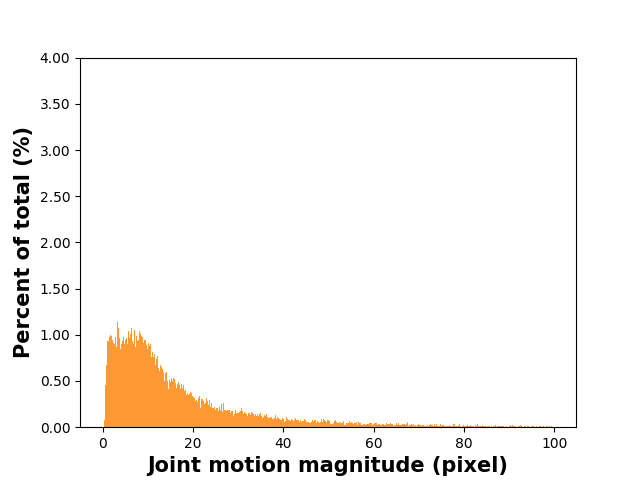}}
    \\
    \caption{\textbf{Statistics on blur strength of the presented BlurHand.}
    On average, the joint motion magnitude from the train and test set are 16.9 and 17.8, respectively.
    }
    \label{fig:motion_blur_strength_statistics}
\end{figure}

In Table~\ref{table:blurhand_detail}, we report the detailed number of training samples.
We note that the right and left hands are evenly distributed in the BlurHand.
In \figref{motion_blur_strength_statistics}, we further report additional measurements, namely joint motion magnitude, to present the statistics on the blur strength of our BlurHand.
%
In detail, we first prepare five sequential sharp frames, which construct a single blurry frame in our BlurHand.
Then we calculate the 2D joint distance between two adjacent sharp frames using the GT joint positions.
Finally, we add all distances for each joint, which we denote as joint motion magnitude.
We note that the large joint motion magnitude means a strong blur exists in hand.
Our BlurHand contains samples with various joint motion magnitude in both the train and test sets.

\section{Results from various deblurring methods}
\label{sec_supp:deblurring}
In Tables~\textcolor{red}{2} and \textcolor{red}{3} in our main manuscript, we compared our BlurHandNet with the combination of state-of-the-art 3D hand mesh estimation methods~[\fakeref{16}, \fakeref{21}, \fakeref{22}] and off-the-shelf deblurring method~[\fakeref{3}].
%
In Table~\ref{sup_table:deblurring}, we additionally compare the results from another widely used deblurring method, DeepDeblur~[\fakeref{25}], as the final mesh estimation results might be dependent on the performance of deblurring methods.
Please note that NAFNet~[\fakeref{3}] is a deblurring method that we used in the main manuscript.
%
Our BlurHandNet still outperforms the case when we use DeepDeblur~[\fakeref{25}] as the deblurring method.
The results again demonstrate that utilizing temporal information is useful rather than simply adopting deblurring methods.

\begin{table}[t]
\small
\centering
\setlength\tabcolsep{1.0pt}
\def\arraystretch{1.1}
\begin{tabular}{C{2cm}|C{2.3cm}C{2.3cm}}
\specialrule{.1em}{.05em}{.05em}
\hline
\multirow{2}{*}{Split} & \multicolumn{2}{c}{BlurHand} \\
& Left hand & Right hand \\
\hline
Train & 85,380 & 83,659 \\
Test & 17,143 & 16,914 \\
\hline
\specialrule{.1em}{.05em}{.05em}
\end{tabular}
\caption{
    \textbf{Number of blurry hand image in our BlurHand.}
    We count both if the image contains both hands.
}
\label{table:blurhand_detail}
\end{table}

\begin{table}[t]
\small
\centering
\setlength\tabcolsep{1.0pt}
\def\arraystretch{1.1}
\resizebox{0.95\linewidth}{!}{
\begin{tabular}{C{2.2cm}|C{1.4cm}|C{1.4cm}|C{1.4cm}|C{1.1cm}|C{1.1cm}|C{1.1cm}}
\specialrule{.1em}{.05em}{.05em}
\hline
Deblurring & \multirow{2}{*}{Deblur} & \multirow{2}{*}{Unfolder} & \multirow{2}{*}{KTFormer} & \multicolumn{3}{c}{MPJPE} \\
\cline{5-7}
methods & & &  & initial & middle & final  \\
\hline
\multirow{2}{*}{DeepDeblur~[\fakeref{25}]} & \cmark & \xmark & \xmark & - & 18.03 & - \\ 
& \cmark & \cmark & \cmark  & 19.24 & 18.04 & 19.27\\
\hline
\multirow{2}{*}{NAFNet~[\fakeref{3}]} & \cmark & \xmark & \xmark & - & 17.28 & - \\ 
& \cmark & \cmark & \cmark & 18.95 & 17.28 & 19.10 \\
\hline
None & \xmark & \cmark & \cmark & \textbf{18.08} & \textbf{16.80} & \textbf{18.21} \\
\hline
\specialrule{.1em}{.05em}{.05em}
\end{tabular}}
\vspace{-3mm}
\caption{
    \textbf{Comparison results on various deblurring methods~[\fakeref{3}, \fakeref{25}]}. Instead of adopting deblurring methods, the proposed BlurHandNet (last row) directly utilizes temporal information from a single blurry image.
}
\label{sup_table:deblurring}
\end{table}

\section{Training details}
\label{sec:training_details}
%

\noindent\textbf{BlurHandNet.}
%
We use Adam optimizer with a batch size of 48 for training our BlurHandNet.
The initial learning rate is set to $1\times 10^{-4}$ and reduced by a factor of $10$ at the $10\textit{th}$ and $12\textit{th}$ epochs.
The proposed network is trained for 13 epochs and takes about 5.7 hours using two NVIDIA 2080 Ti GPUs.
All other details will be available in our codes.

\noindent\textbf{State-of-the-art models.}
%
For training the state-of-the-art 3D hand mesh estimation networks~[\fakeref{16}, \fakeref{21}, \fakeref{22}] and deblurring networks~[\fakeref{3}] on BlurHand, we follow their official training instruction.
%
In addition, we employ the authors' official pre-trained weight in training deblurring methods~[\fakeref{3}] for easier optimization
\section{Additional qualitative results}
\label{sec:more_results}

\begin{figure*}[t]
    \centering
    \newcommand{\spacing}{3.3cm}
    \captionsetup[subfigure]{labelfont=scriptsize, textfont=scriptsize}
        \subfloat{\includegraphics[width=\spacing]{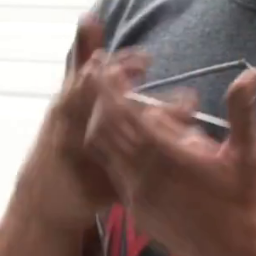}}
        \hfill
        \subfloat{\includegraphics[width=\spacing]{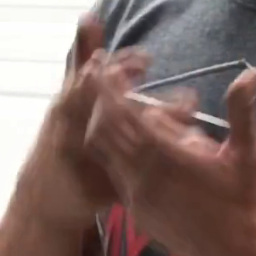}}
        \hfill
        \subfloat{\includegraphics[width=\spacing]{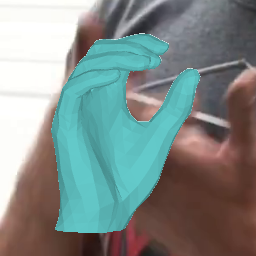}}
        \hfill
        \subfloat{\includegraphics[width=\spacing]{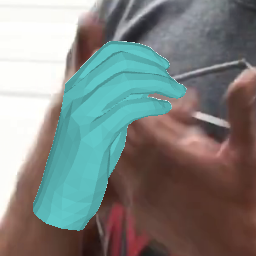}}
        \hfill
        \subfloat{\includegraphics[width=\spacing]{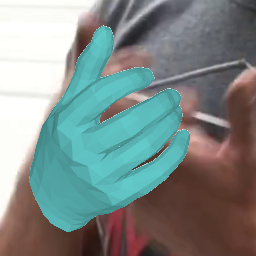}}
    \\
    \vspace{1mm}
        \subfloat{\includegraphics[width=\spacing]{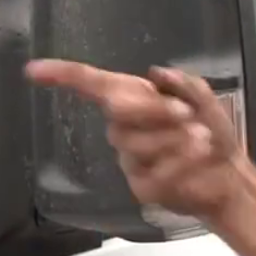}}
        \hfill
        \subfloat{\includegraphics[width=\spacing]{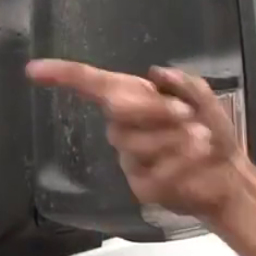}}
        \hfill
        \subfloat{\includegraphics[width=\spacing]{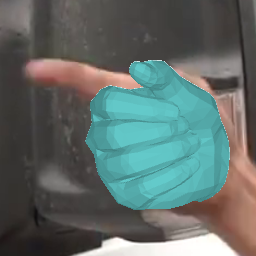}}
        \hfill
        \subfloat{\includegraphics[width=\spacing]{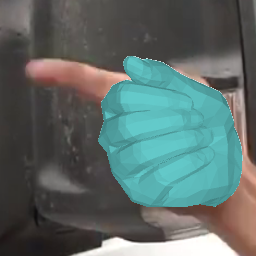}}
        \hfill
        \subfloat{\includegraphics[width=\spacing]{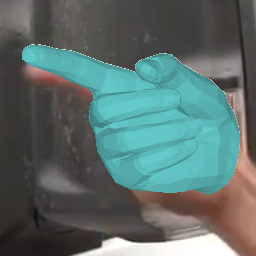}}
    \\
    \vspace{1mm}
        \subfloat{\includegraphics[width=\spacing]{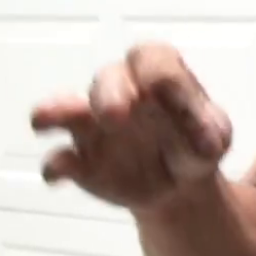}}
        \hfill
        \subfloat{\includegraphics[width=\spacing]{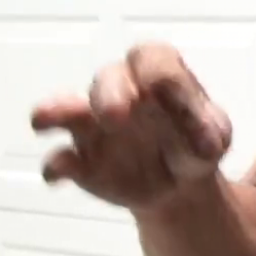}}
        \hfill
        \subfloat{\includegraphics[width=\spacing]{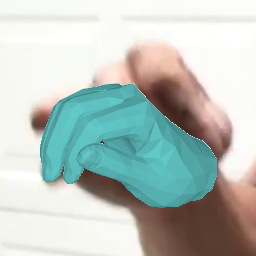}}
        \hfill
        \subfloat{\includegraphics[width=\spacing]{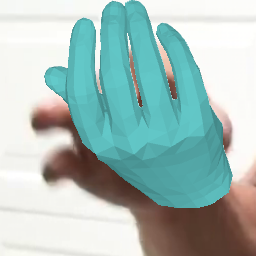}}
        \hfill
        \subfloat{\includegraphics[width=\spacing]{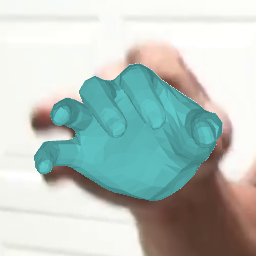}}
    \\
    \vspace{1mm}
        \subfloat{\includegraphics[width=\spacing]{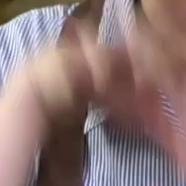}}
        \hfill
        \subfloat{\includegraphics[width=\spacing]{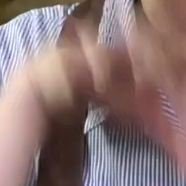}}
        \hfill
        \subfloat{\includegraphics[width=\spacing]{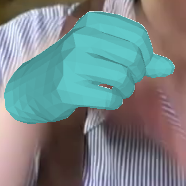}}
        \hfill
        \subfloat{\includegraphics[width=\spacing]{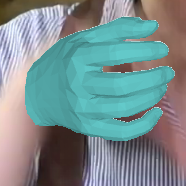}}
        \hfill
        \subfloat{\includegraphics[width=\spacing]{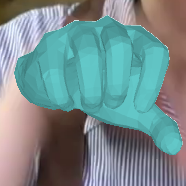}}
    \\
    \vspace{1mm}
        \subfloat{\includegraphics[width=\spacing]{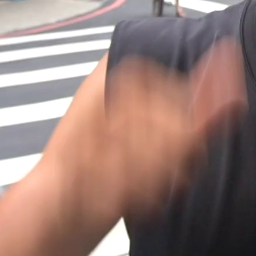}}
        \hfill
        \subfloat{\includegraphics[width=\spacing]{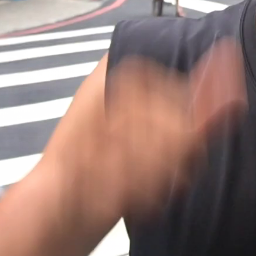}}
        \hfill
        \subfloat{\includegraphics[width=\spacing]{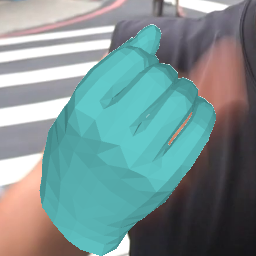}}
        \hfill
        \subfloat{\includegraphics[width=\spacing]{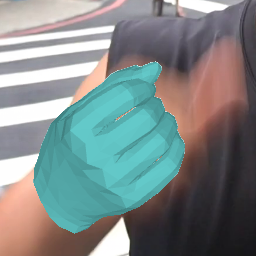}}
        \hfill
        \subfloat{\includegraphics[width=\spacing]{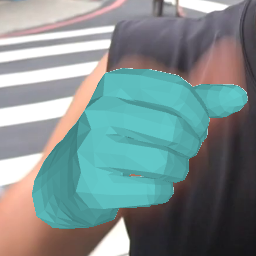}}
    \\
    \addtocounter{subfigure}{-25}
        \subfloat[Blurry image]{\includegraphics[width=\spacing]{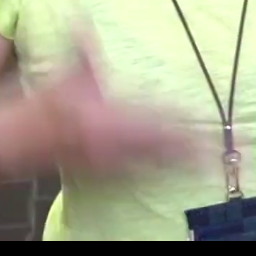}}
        \hfill
        \subfloat[Deblurred image]{\includegraphics[width=\spacing]{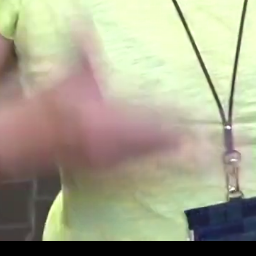}}
        \hfill
        \subfloat[IH2.6M + YT3D]{\includegraphics[width=\spacing]{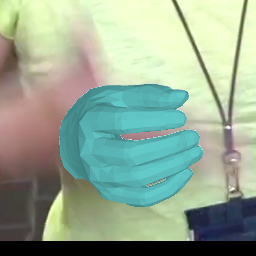}}
        \hfill
        \subfloat[BH+\textbf{D} + YT3D]{\includegraphics[width=\spacing]{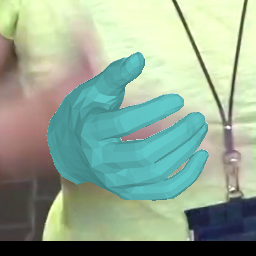}}
        \hfill
        \subfloat[BH + YT3D]{\includegraphics[width=\spacing]{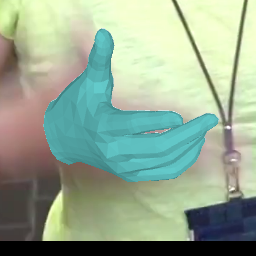}}
    \caption{\textbf{Effectiveness of the presented BlurHand.}
    The captions below figures describe training sets used to train 3D hand mesh estimation networks.
    The notation \textbf{D} represents that the network is trained on deblurred BH and tested on (b).
    }
    \vspace{-0.472cm}
    \label{fig:supple_comparison_yt3d0}
\end{figure*}

\noindent\textbf{Effectiveness of the BlurHand.}
In Figure~\ref{fig:supple_comparison_yt3d0}, we provide additional qualitative results on YT-3D~[\fakeref{13}].
We note that training the model on our BlurHand~(column (e) in the \figref{supple_comparison_yt3d0}) is significantly helpful in dealing with the in-the-wild blurry hand images compared to the cases using sharp images and deblurred images~(column (c) and (d) in the \figref{supple_comparison_yt3d0}).
The results justify the necessity of our BlurHand when handling the blurry hand.

\begin{figure*}[t]
\begin{center}
\includegraphics[width=1.0\linewidth]{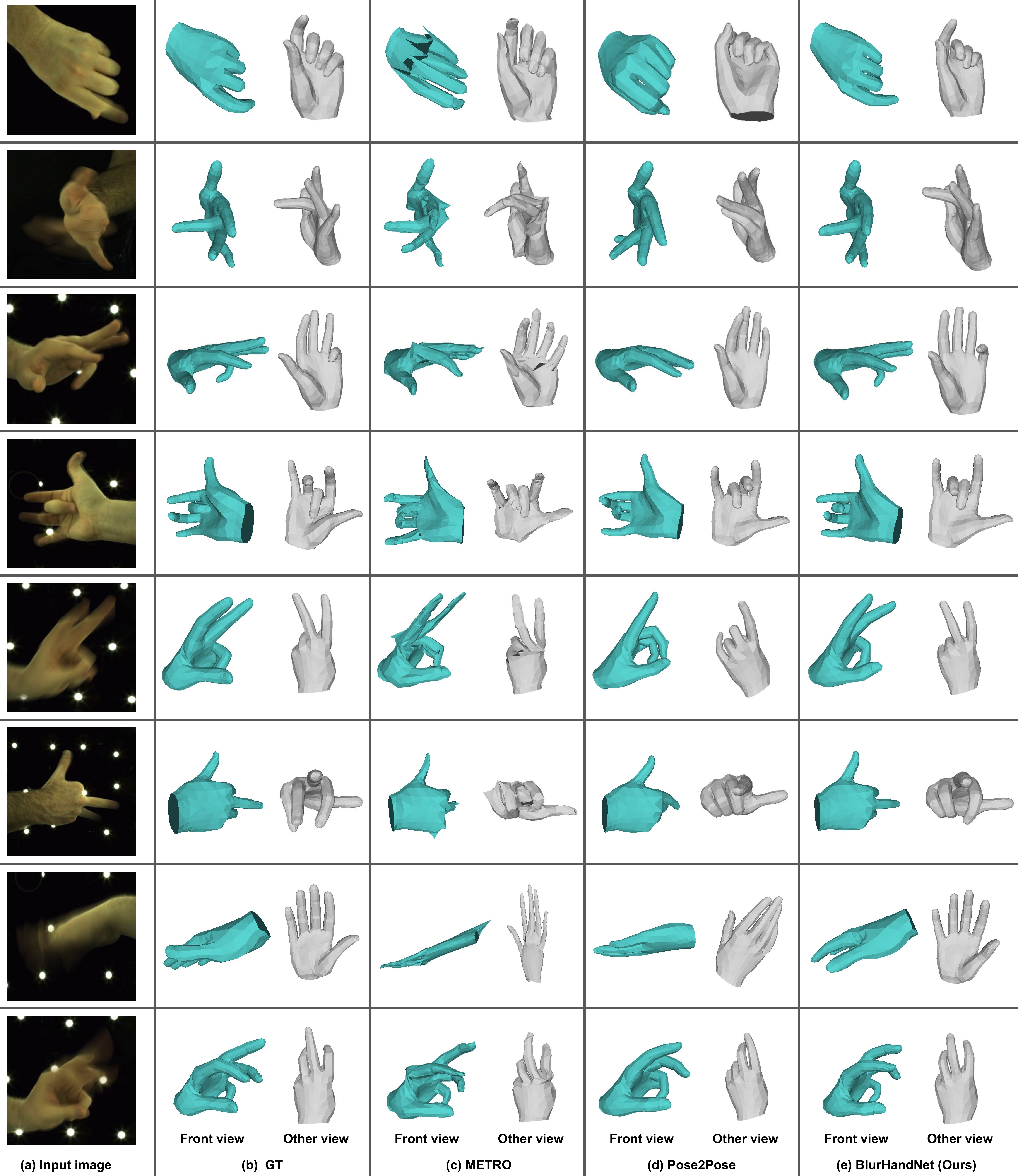}
\end{center}
\vspace*{-5mm}
\caption{\textbf{Visual comparison of the proposed BlurHandNet and state-of-the-art 3D hand mesh estimation methods~[\fakeref{16}, \fakeref{21}] on BlurHand.}
We note that all the methods are trained on BlurHand.
}
\label{fig:comparison_BlurHand}
\vspace*{-6mm}
\end{figure*}

\noindent\textbf{Visual comparison on BlurHand.}
\label{ssec:limitations}
In \figref{comparison_BlurHand}, we present additional comparison results on BlurHand.
%
Compared to the previous state-of-the-art methods~[\fakeref{16}, \fakeref{21}], our BlurHandNet reconstructs more accurate 3D hand meshes by exploiting temporal information.
%


\section{Discussion}
\label{sec:discussion}

\begin{figure}[!t]
    \newcommand{\ww}{0.44}
    \centering
    \newcommand{\spacing}{3.3cm}
    \subfloat{\includegraphics[width=\spacing]{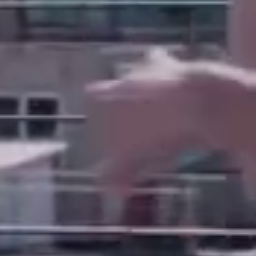}}
    \hspace{4mm}
    \subfloat{\includegraphics[width=\spacing]{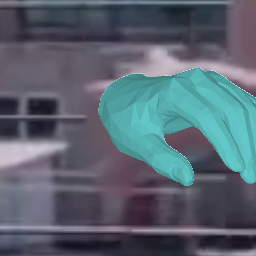}}
    \addtocounter{subfigure}{-2}
    \\
    \vspace{1mm}
    \subfloat[Input image]{\includegraphics[width=\spacing]{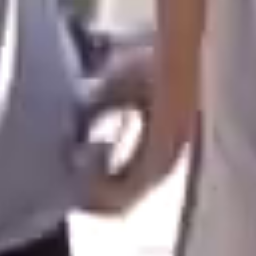}}
    \hspace{4mm}
    \subfloat[BlurHandNet (Ours)]{\includegraphics[width=\spacing]{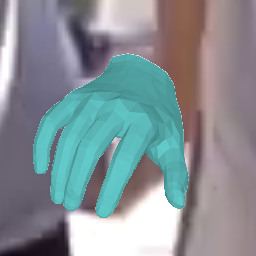}}
    \vspace{-2mm}
    \captionsetup[subfigure]{labelformat=empty}
    \begin{minipage}{1.0\linewidth}
    \vspace{1.5mm}
    \subcaption{(blurry and low-resolution) \textcolor{white}{AAAAAAAAAAAAAAAAAA}}
    \end{minipage}
    \caption{\textbf{Failure cases.}
    Our BlurHandNet produces less accurate results when inputs contain multiple complex degradations.
    }
    \label{fig:failure_case}
\end{figure}

\noindent\textbf{Limitations and future works.}
While various types of image degradations are prevalent in real-world hand images, \eg, low-resolution, noise, and low illumination, we especially focus on hands with blur artifacts.
%
Figure~\ref{fig:failure_case} shows that BlurHandNet produces not robust results when the input image is low-resolution with blur, which should be considered in our future works.

\noindent\textbf{Societal impacts.}
Our BlurHand and BlurHandNet suggest a new and necessary research direction toward real-world applications, the robust 3D hand mesh estimation from blurry images.
In particular, our method can be useful for AR/VR as people often move hands fast, which causes motion blur of hands.

